\documentclass{article}
\usepackage{arxiv}

\usepackage[utf8]{inputenc} 
\usepackage[T1]{fontenc}    
\usepackage{hyperref}       
\usepackage{url}            
\usepackage{booktabs}       
\usepackage{amsfonts}       
\usepackage{nicefrac}       
\usepackage{microtype}      
\usepackage{lipsum}		
\usepackage{graphicx}
\usepackage{natbib}
\usepackage{doi}

\usepackage{amsmath,amsfonts}
\usepackage{algorithmic}
\usepackage{array}
\usepackage[caption=false,font=normalsize,labelfont=sf,textfont=sf]{subfig}
\usepackage{threeparttable}
\usepackage{textcomp}
\usepackage{stfloats}
\usepackage{url}
\usepackage{verbatim}
\usepackage{multirow}
\usepackage{graphicx}
\usepackage{blindtext}
\usepackage[T1]{fontenc}
\usepackage{lscape}
\usepackage{comment}
\usepackage{xr-hyper}
\usepackage[table,xcdraw]{xcolor}
\usepackage{enumitem}
\usepackage{tcolorbox}

\usepackage{amsmath,amsfonts,amssymb,graphicx}
\usepackage{subfiles}
\usepackage{hyperref} 

\title{One Size Does Not Fit All: Exploring Variable Thresholds for Distance-Based Multi-Label Text Classification}

\author{{\hspace{1mm}Jens Van Nooten}\thanks{These authors contributed equally to this work.} \\
	University of Antwerp (CLiPS)\\
	\And
	{\hspace{1mm}Andriy Kosar$^{*}$} \\
	Textgain\\
    \And
	{\hspace{1mm}Guy De Pauw} \\
	Textgain\\
    \And
	{\hspace{1mm}Walter Daelemans} \\ 
    University of Antwerp (CLiPS)\\
}

\date{}

\renewcommand{\headeright}{}
\renewcommand{\undertitle}{}
\renewcommand{\shorttitle}{One Size Does Not Fit All}

\hypersetup{
pdftitle={One Size Does Not Fit All: Exploring Variable Thresholds for Distance-Based Multi-Label Text Classification},
pdfsubject={cs.AI, cs.LG, cs.CL},
pdfauthor={Jens Van Nooten, Andriy Kosar, Guy De Pauw, Walter Daelemans},
pdfkeywords={multi-label classification, semantic similarity, text classification},
}

\begin{document}
\maketitle

\begin{abstract}
Distance-based unsupervised text classification is a method within text classification that leverages the semantic similarity between a label and a text to determine label relevance. This method provides numerous benefits, including fast inference and adaptability to expanding label sets, as opposed to zero-shot, few-shot, and fine-tuned neural networks that require re-training in such cases. In multi-label distance-based classification and information retrieval algorithms, thresholds are required to determine whether a text instance is ``similar'' to a label or query. Similarity between a text and label is determined in a dense embedding space, usually generated by state-of-the-art sentence encoders.
Multi-label classification complicates matters, as a text instance can have multiple true labels, unlike in multi-class or binary classification, where each instance is assigned only one label. We expand upon previous literature on this underexplored topic by thoroughly examining and evaluating the ability of sentence encoders to perform distance-based classification. First, we perform an exploratory study to verify whether the semantic relationships between texts and labels vary across models, datasets, and label sets by conducting experiments on a diverse collection of realistic multi-label text classification (MLTC) datasets. We find that similarity distributions show statistically significant differences across models, datasets and even label sets. We propose a novel method for optimizing label-specific thresholds using a validation set. Our label-specific thresholding method achieves an average improvement of 46\% over normalized 0.5 thresholding and outperforms uniform thresholding approaches from previous work by an average of 14\%. Additionally, the method demonstrates strong performance even with limited labeled examples.
\end{abstract}


\section{Introduction}
Multi-label text classification (MLTC) is a text classification problem where the goal is to predict one or multiple labels from a given finite label set for a single text. This type of classification is more challenging than multi-class classification, where only one label is assigned to each item, because it involves predicting multiple labels simultaneously. This introduces complexities such as an exponentially growing number of label combinations, interdependence between labels, and the fact that the number of labels for each instance can vary~\cite{tarekegn2024deeplearningmultilabellearning}. Furthermore, this entails that the semantics of multiple labels are represented in a single text instance.

Even though this classification problem is more complex and difficult to solve than single-label classification, it has found many real-world applications with text data, such as topic classification in the news sector~\cite{Isnaini_2019} and commerce~\cite{axioms11090436}, emotion classification~\cite{deng-fuji-2023-multilabel}, social media monitoring~\cite{huang-etal-2013-social, Lemmens_Dejaeghere_Kreutz_VanNooten_Markov_Daelemans_2021, vannootendaelemans2023improving}, and COVID-19 symptom prediction~\cite{van-olemn-etal-2022-predicting}. Additionally, it is often employed in image \cite{xiongyanmei2018subject, wangetal2019baseline}, video~\cite{gupta2024openvocabularymultilabelvideo}, and audio~\cite{mulimani2024classincrementallearningmultilabelaudio} classification. Other examples of applications include the classification of proteins~\cite{forero-etal-2013-comparison, szalkaigrolmusz2018protein}. 

Distance-based classification (DBC) is a classification method that leverages the semantic similarity between texts and labels in a common embedding space to determine whether a label belongs to a text or not~\cite{chang2008importance}. In traditional distance-based classification for binary or multi-class classification, the most common approach is to assume that the closest label embedding to a text embedding is the true label~\cite{Kosar-etal-2022, kosar-etal-2023-advancing}. However, as previously mentioned, the number of labels to be predicted in multi-label classification is unknown. Therefore, an algorithm to determine the number of labels should be introduced. In distance-based classification and information retrieval, this is often achieved by setting an appropriate threshold. The main question in distance-based MLTC therefore is a question of how to optimally determine similarity thresholds to optimize performance. Although some work has explored DBC for MLTC~\cite{veeranna2016using,mylonas-etal-2020-zero,mustafa2021multi,sarkaretal2023zero}, most proposed threshold algorithms only consider a single (uniform) threshold for all labels, ignoring differences in similarity distributions between texts and individual labels. We argue that label-specific thresholds, which account for these differences, yield significantly improved performance on downstream multi-label classification tasks.

This study consists of two parts. First, using multiple MLTC datasets and sentence encoders, it examines similarity scales, i.e. the range between the lower bound and upper bound expressed by the model when measuring the similarity between labels and texts in an embedding space. Second, based on the results of this exploratory analysis, it investigates the effectiveness and feasibility of implementing label-specific thresholds for MLTC. 

The contributions of this empirical study are the following: 
\begin{enumerate}
    \item  We investigate an under-explored approach to multi-label text classification using a distance-based classification method, offering an efficient alternative to more commonly used methods.
    \item We demonstrate that uniform thresholds are suboptimal for multi-label classification tasks, and that label-specific thresholds lead to superior performance.
    \item We propose a simple, novel and intuitive method for determining label-specific thresholds with minimal annotated data. Our method significantly outperforms existing uniform thresholding techniques and achieves competitive results compared to state-of-the-art zero-shot classification with LLMs across several publicly available datasets.
    \item  We conduct a comprehensive evaluation of open-source state-of-the-art sentence encoders for multi-label classification, assessing their effectiveness in addressing complex classification challenges.
\end{enumerate}

The findings of this study are not only applicable to MLTC but also have potential applications in information retrieval, particularly in Retrieval-Augmented Generation (RAG). Moreover, our findings could be extended to other modalities, such as video, audio, and images.

The paper is structured as follows. Section~\ref{sec:problem} defines the problem and presents the key hypotheses of this empirical study. Section~\ref{sec:related-research} reviews relevant studies that form the foundation of the proposed methodology. Subsequently, Section~\ref{sec:methodology} outlines the methodology: It begins with the exploratory study, followed by an explanation of the classification experiments. Then, Section~\ref{sec:results} presents the results and analysis of the experimental study and the classification experiments.

\section{Problem Statement and Hypotheses} \label{sec:problem}

We formulate the problem as a threshold optimization problem, following prior work~\cite{veeranna2016using, mylonas-etal-2020-zero, mustafa2021multi, sarkar-etal-2022-exploring, sarkaretal2023zero}. We build on previous literature by finding the threshold for each label individually to maximize the overall performance. Let $L = \{l_1, l_2 ... l_n\}$ be the set of labels and let $\theta = \{0.01, 0.02, 0.03 ... 1.0\}$ be a set of pre-defined thresholds $\theta$. The goal of the proposed algorithm is to find a threshold for each label ($\theta^*_{L}$) that yields the highest performance ($F_1$ score): 

\begin{equation}
\theta^*_{L} = arg\ max\ F_{1_l}(\theta)
\end{equation}

We formulate the following four key hypotheses:

\begin{enumerate}
    \item \textbf{Variability in Embedding Scales Across Models (H1)}: Different embedding models exhibit unique similarity scales, which raises challenges for their interchangeable application. 
    \item \textbf{Domain-Specific Differences (H2)}: Even within a single model, texts from various genres or domains can demonstrate distinct similarity scales, limiting cross-domain applications.
    \item \textbf{Class-Specific Similarity Variations (H3)}: Within the semantic space of a single model and domain, different classes may exhibit unique similarity scales. This variation depends on how effectively the embedding model captures and reflects the semantic meaning of texts and labels, particularly for labels with weak representations.
    \item \textbf{Class-Specific Thresholds (H4)}: Optimized class-specific thresholds enhance the performance of multi-label distance-based text classification.
\end{enumerate}

\section{Related Research}\label{sec:related-research} 
\subsection{Multi-Label Text Classification}
As mentioned previously, multi-label classification carries unique difficulties as opposed to single-label classification. Researchers have previously attempted to overcome these challenges by transforming the problem into multiple binary classification problems~\cite{zhang-etal-2018-binaryrelevance}, modeling hierarchies~\cite{fan-qiu-2023-hierarchical} and label correlations \cite{jia-etal-2022-research}, performing data augmentation~\cite{vannootendaelemans2023improving, cai-etal-2023-resolving}, or adapting loss functions~\cite{benbaruch2021asymmetricloss}. For a more comprehensive overview of multi-label classification approaches and challenges, consult~\cite{tarekegn2024deeplearningmultilabellearning} and~\cite{li-etal-2022-survey}.  

Recent advances in large language models (LLMs), such as BERT~\cite{devlin-etal-2019-bert} and the GPT series~\cite{Radford2018ImprovingLU}, have transformed MLTC --and text classification in general-- into a zero-shot or few-shot task~\cite{yin2024crisissensellminstructionfinetunedlarge, malik-etal-2024-pseudo, van-nooten-kosar-2024-advancing}. However, as highlighted in~\cite{wang-etal-2023-text2topic} and~\cite{van-nooten-kosar-2024-advancing}, state-of-the-art generative LLMs still perform poorly on MLTC tasks, mainly due to the complexity of the task and the limited availability of official annotation guidelines that can be included in prompts. Nevertheless, deploying LLMs for MLTC in an industrial setting is attractive because they do not require task-specific training data and enable flexibility, especially with regard to classification on datasets with evolving label sets. Compared to fine-tuned models, they do not need to be retrained whenever a label is added or removed from a dataset. However, as highlighted in~\cite{BOGATINOVSKI2022117215}, many multi-label classification approaches may suffer from high computational complexity, which makes them less applicable in industrial settings. In such cases, models are required to be lightweight and fast. 

\subsection{Distance-based Text Classification}
Alongside supervised learning, distance-based classification (DBC) has evolved as a computationally efficient method, where the similarity between text and label embeddings is used to determine whether a label belongs to a text or not. Labels and texts are usually encoded using neural word embeddings such as \textit{GloVe}~\cite{pennington-etal-2014-glove}, \textit{word2vec}~\cite{word2vec} or \textit{fastText}~\cite{fasttext}, or using contextual embeddings with sentence transformers~\cite{reimers2019sentence-bert}. This method was introduced by~\cite{chang2008importance}, who encoded texts and semantic class concepts from Wikipedia as labels in a semantic space to perform classification based on the cosine similarity between label and text. In single-label text classification, as performed in~\cite{kosar-etal-2023-advancing}, usually the closest label is considered to be the true label. In this approach, both labels and texts only have to be encoded once, which makes it lightweight and fast compared to inference with LLMs, since inference only requires a similarity metric to be calculated.

However, this becomes more complex when performing distance-based MLTC, since there are multiple potential true label candidates for a single text instance. Encoding the multiple crucial facets of a multi-label text effectively remains a challenge. Several studies have addressed this complexity using distance-based methods for MLTC~\cite{veeranna2016using, mylonas-etal-2020-zero, mustafa2021multi, sarkar-etal-2022-exploring, sarkaretal2023zero}. In these approaches, both texts and label representations are also embedded in a joint space, after which the similarity between a text embedding and label embedding is calculated. If the similarity (typically cosine similarity) equals or exceeds a threshold, a label is assigned to the text.

Although there are several ways to determine these thresholds, most research opts for using user-defined thresholds, selecting the best-performing one based on validation set performance~\cite{veeranna2016using, sarkar-etal-2022-exploring, sarkaretal2023zero}. However, this method ignores the differences in similarity scales between labels and texts, because the same threshold is applied to every label. Not optimizing the threshold for each label separately when a validation set is available could inhibit performance. Table~\ref{tab:diff-sims} presents text examples that highlight the need for label-specific thresholds, as illustrated by the variation in cosine similarity values between both models and labels.

\begin{table*}[ht]

\centering
\caption{Comparison of cosine similarity values between a text and the corresponding true labels across models}
\label{tab:diff-sims}
\resizebox{1\textwidth}{!}{%
\begin{tabular}{l|cc|cccc}
 & \multicolumn{2}{c|}{\textbf{LitCovid}} & \multicolumn{4}{c}{\textbf{Reuters}} \\ \hline
\textbf{Text} & \multicolumn{2}{l|}{\begin{tabular}[c]{@{}l@{}}In this study, we aimed to assess the association between \\ development of cardiac injury and short-term mortality as \\ well as poor in-hospital outcomes in hospitalized \\ patients with COVID-19.\end{tabular}} & \multicolumn{4}{l}{\begin{tabular}[c]{@{}l@{}}Soybean imports are forecast to rise to 425,000 tonnes in 1987/88 \\ (October/September) from an estimated 300,000 in 1986/87 and \\ 375,000 in 1985/86, the U.S. Embassy said in its annual report \\ on Indonesia's agriculture.\end{tabular}} \\
 &  &  &  &  &  &  \\
\textbf{True Labels} & Medical Treatments & Diagnostic Methods & Oilseed & Soybeans & Meal and Feed & Soybean Meal \\ \hline
 &  &  &  &  &  &  \\
\textbf{GIST-Large} & 0.31 & 0.27 & 0.44 & 0.65 & 0.32 & 0.52 \\
\textbf{Stella} & 0.28 & 0.30 & 0.41 & 0.58 & 0.36 & 0.58 \\
\textbf{Mxbai} & 0.43 & 0.40 & 0.51 & 0.66 & 0.43 & 0.56
\end{tabular}%
}
\end{table*}

Determining optimal dynamic thresholds has also been researched for RAG. For example, the study by~\cite{radeva-etal-2024-similarity} highlights that different models require distinct similarity thresholds to reach optimal performance in RAG due to variations in model architectures, interpretations of similarity scores, and training procedures. However, the authors do not explore variations in similarity by domain or specific query type. In the present study, we expand on this optimization process by accounting for variability in the encoded semantics of each label in a multi-label context.

\section{Methodology}\label{sec:methodology}
This study is divided into two parts, for which the methodology is described in the following sections. In the first part, we explore the aforementioned hypotheses, i.e. the variability in similarity distributions among models~(\textbf{H1}), datasets~(\textbf{H2}), and labels~(\textbf{H3}) by conducting statistical analyses on similarity distributions derived from the labeled training partitions of the datasets. The second part aims to assess the feasibility of label-specific thresholds by applying them to a broad range of MLTC datasets~(\textbf{H4}). 

\subsection{Datasets}
The experiments are conducted on five publicly available MLTC datasets, namely: the English subset of the SemEval 2018 dataset~\cite{mohammad-etal-2018-semeval}, the BioTech news dataset\footnote{https://blog.knowledgator.com/finally-a-decent-multi-label-classification-benchmark-is-created-a-prominent-zero-shot-dataset-4d90c9e1c718}, a modified version ("Apté Mod") of the Reuters-21578 dataset~\cite{apte-etal-reuters}\footnote{https://huggingface.co/datasets/ucirvine/reuters21578}, the arXiv Academic Papers Dataset (AAPD)~\cite{yang-etal-2018-sgm}\footnote{For the AAPD dataset, we mapped the abbreviations to the descriptive names on the arXiv website. When an abbreviation mapped to multiple possible label names, we added the general domain in brackets.} and the LitCovid dataset~\cite{chen-etal-litcovid}. An overview of the datasets is provided in Table~\ref{tab:datasets}. These datasets were selected based on the availability of the original text data and the inclusion of label names, which are essential for distance-based text classification. For the AAPD and LitCovid datasets, we created a validation split by performing a stratified split on the training data, following the method described in~\cite{mlstratification}.

\begin{table}[t!]
\caption{Dataset Statistics}
\centering
\resizebox{0.80\columnwidth}{!}{%
\begin{tabular}{l|ccccccccc}
\textbf{Dataset} & \textbf{\begin{tabular}[c]{@{}l@{}}Text \\ Type\end{tabular}} & Task & \textbf{N Train} & \textbf{N Val} & \textbf{N Test} & \textbf{N Lbls} & \textbf{\begin{tabular}[c]{@{}l@{}}N lbls/ \\ Text\end{tabular}} & \textbf{\begin{tabular}[c]{@{}l@{}}Mn \\ Tkns\end{tabular}} & \textbf{\begin{tabular}[c]{@{}l@{}}Mdn \\ Tkns\end{tabular}} \\ \hline
\textbf{SemEval} & tweets & emotion & 6,837 & 886 & 3,259 & 11 & 2.38 & 27 & 28 \\
\textbf{BioTech} & news & topics & 2,344 & 414 & 381 & 31 & 1.84  & 655 & 572  \\
\textbf{Reuters} & news & topics & 7,493 & 1,323 & 3,023 & 118 & 1.01 & 180 & 121 \\
\textbf{AAPD} & abstracts & topics & 53,840 & 1,000 & 1,000 & 52 & 2.41 & 128 & 99 \\
\textbf{LitCOVID} & abstracts & topics & 21,204 & 3,756 & 6,239 & 7 & 1.37 & 303 & 292
\label{tab:datasets}
\end{tabular}%
} 
\end{table}


\subsection{Exploratory Study}
For the exploratory study, we first embed the texts from the training partitions and all corresponding labels of the data using neural word embeddings such as \textit{GloVe}~\cite{pennington-etal-2014-glove} and sentence transformers~\cite{reimers2019sentence-bert}. The latter models are selected based on their ranking on the MTEB leaderboard\footnote{https://huggingface.co/spaces/mteb/leaderboard}, their size (in number of parameters), and their public availability. To reduce computational cost, we ensured that all experiments could be run on a single \textit{NVIDIA GeForce RTX 2080 Ti} GPU\footnote{Table~\ref{tab:times} in Appendix~\ref{app:times} shows the computation times for each method.}. Table~\ref{tab:models} provides an overview of all the embedding models used in the experiments.

\begin{table}[t]
\caption{Overview of Embedding Models\protect\footnotemark}
\centering
\resizebox{0.70\columnwidth}{!}{%
\begin{tabular}{lccc}
\textbf{Model} & \multicolumn{1}{l}{\textbf{Size (M)}} & \multicolumn{1}{l}{\textbf{Embedding Dim.}} & \multicolumn{1}{l}{\textbf{Max Tokens}} \\ \hline
(1)~Gist-Large~\cite{solatorio2024gistembed} & 335 & 1024 & 512 \\
(2)~GTE-Large~\cite{li2023towards} & 434 & 1024 & 8192 \\
(3)~Stella & 435 & 8192 & 512 \\
(4)~UAE-Large~\cite{li2023angle} & 335 & 1024 & 512 \\
(5)~Mxbai~\cite{emb2024mxbai} & 335 & 1024 & 512 \\
(6)~BGE & 109 & 768 & 512 \\
(7)~SF & 335 & 1024 & 512 \\
(8)~all-MPNet & 109 & 768 & 512 \\
(9)~all-MPNet-Wiki~\cite{kosar-etal-2023-advancing} & 109 & 768 & 512 \\
(10)~GloVe~\cite{pennington-etal-2014-glove} &  & 300 & 40,001
\end{tabular}%
}
\label{tab:models}
\end{table}
\footnotetext{
Model details: \\
(1)~\url{https://huggingface.co/gist-large-embedding-v0} \\
(2)~\url{https://huggingface.co/Alibaba-NLP/gte-large-en-v1.5} \\
(3)~\url{https://huggingface.co/dunzhang/stella_en_400M_v5} \\
(4)~\url{https://huggingface.co/WhereIsAI/UAE-Large-V1} \\
(5)~\url{https://huggingface.co/mxbai/Embed-Large-v1} \\
(6)~\url{https://huggingface.co/BAAI/bge-base-en-v1.5} \\
(7)~\url{https://huggingface.co/jamesgpt1/sf_model_e5} \\
(8)~\url{https://huggingface.co/sentence-transformers/all-mpnet-base-v2} \\
(9)~\url{https://huggingface.co/textgain/TopicAwareSTallmpnetbasev2Wiki} \\
(10) glove.6B.300d, \url{https://nlp.stanford.edu/projects/glove/}. Token limit per \textit{sentence-transformers}.
}

Both label names and texts from the training sets are encoded using these models to obtain their embeddings. Given the text embeddings and the embeddings of all labels, we then calculate the cosine similarity between the text embeddings and all label embeddings to obtain distribution $\theta$. We then split~$\theta$ into two sub-distributions: $\alpha$ and $\beta$. Here, $\alpha_{ij}$ represents the similarity score between the text embedding with index~$i$ and the corresponding label with index $j$ that is part of the true label set~$\mathbf{y}_i$. On the other hand, $\beta_{ik}$ represents the similarity scores between the text embedding with index~$i$ and a label with index~$k$ that is not part of~$\mathbf{y}_i$.

To compare similarity distributions and highlight variability between models~(\textbf{H1}) and domain-specific variability~(\textbf{H2}), we normalize the cosine similarities (cf. Figure~\ref{fig:domain_variability}) with min-max normalization (cf. Appendix~\ref{app:hyp1}, Figure~\ref{fig:norm_domain_variability}). Given $\theta$, a vector containing cosine similarity scores, we obtain $\theta'$ as follows:

\[
\mathbf{\theta}' = \frac{\mathbf{\theta} - \min(\mathbf{\theta})}{\max(\mathbf{\theta}) - \min(\mathbf{\theta})}
\]

In accordance with the first three hypotheses addressed previously, we conduct t-tests with Bonferroni correction between:
\begin{enumerate}
    \item Similarity distributions $\theta$ from each model, done in a pairwise fashion for each possible pair of models. This way, we can test the hypothesis that different embedding models exhibit unique similarity scales (\textbf{H1)}.
    \item Similarity distributions $\theta$ from different datasets, but from the same model. This supports the hypothesis that, within a single model, texts from various genres or domains can demonstrate distinct similarity scales (\textbf{H2)}.
    \item Similarity distributions in $\alpha$ from the same dataset and model, for each possible label pair in a dataset. Thus, we are able to verify the hypothesis that different classes may exhibit unique similarity scales within the semantic space of a single model and dataset (\textbf{H3)}.
\end{enumerate}

For \textbf{H1}, we report the average mean and median cosine similarity percentage of statistically significant ($p<.05$) differences between the cosine similarity distributions of label pairs in a dataset, per model. For \textbf{H2}, we report the proportion of dataset pairs with statistically significant differences in their similarity distributions. Finally, for \textbf{H3}, we report the proportion of label pairs that exhibit statistically significant differences in similarity distributions.

\subsection{Distance-based MLTC} \label{sec:dbc}
\subsubsection{Text Embeddings}
Similarly to the exploratory study, we first embed texts and label names using sentence transformers, following the original implementation of BERT-based sentence encoders in~\cite{reimers2019sentence-bert}. We use the appropriate pooling operation for each model to obtain embeddings. This yields a text embedding batch~$T$. Let~\( T = \{ \mathbf{t}_1, \mathbf{t}_2, \dots, \mathbf{t}_m \} \) denote the set of text embeddings, where~\( \mathbf{t}_i \in \mathbb{R}^d \) represents the embedding of the~\(i\)-th text. $T$~is a tensor with shape~($m$, $h$), where~$m$ is the number of texts in the dataset, and~$h$ is the embedding dimension of a model~$E$. 

\subsubsection{Label Embeddings}
To obtain label embeddings, we explore various methods of representing labels, including standard (original) label names and enhanced representations that improve semantic clarity:
\begin{enumerate}
    \item \textbf{Label Names}: Each label name is used as is from the original dataset. We embed the label names and obtain the label embedding set $L$. 
    \item \textbf{Adjusted Label Names}: In some cases, the label names are not semantically distinct enough\footnote{For example, ``Mechanism'' in the LitCovid dataset}. To address this, we manually adjust the names to make them more distinct and semantically meaningful within the dataset. We embed the adjusted label names and obtain the label embeddings $L$. It should be noted that the label names of the SemEval dataset and AAPD dataset were retained without modification\footnote{We prepended each emotion label in SemEval with \textit{Emotional State:}, but this did not yield any improvements.}.
    \item \textbf{Averaged Keyword Embeddings}: For each (adjusted) label name, we generate 10 additional keywords that are related to the label name. In contrast to~\cite{sarkaretal2023zero}, we employ GPT-4o. We obtain embeddings for each keyword using sentence encoder $E$ and average all embeddings, including the label name embedding, resulting in the label embedding set $L$. With this approach, we aim to expand the search spaces for the models without causing overlap between labels in the embedding space.
\end{enumerate}

Let \( L = \{ \mathbf{l}_1, \mathbf{l}_2, \dots, \mathbf{l}_n \} \) be the set of label embeddings, where \( \mathbf{l}_j \in \mathbb{R}^d \) represents the embedding of the \( j \)-th label. $L$~is a tensor with shape~$(n, h)$, where $n$~is the number of labels in the dataset and $h$~is the hidden size of embedding model~$E$. Examples of each label representation method can be found in Table~\ref{tab:label-representations}. 

\begin{table}[t]
\caption{Label Representation Methods}
\centering
\resizebox{0.70\columnwidth}{!}{%
\begin{tabular}{l|l}
\textbf{Representation} & \textbf{Example} \\ \hline
Label Name & Mechanism \\
Adjusted Label Name & Biological Mechanisms \\
Keywords & \begin{tabular}[c]{@{}l@{}}``Viral Replication'', ``Pathogenesis'', ``Immune Response'', \\ ``Viral Entry'', ``Host Interaction'', ``Molecular Biology'', \\ ``Virus Lifecycle'', ``Cellular Mechanism'', \\ ``Infection Mechanism'', ``Viral Transmission''\end{tabular}
\end{tabular}%
}
\label{tab:label-representations}
\end{table}

\subsubsection{Similarity Calculation and Thresholding}
Given the text embeddings $T$ and label embeddings $L$, we calculate the pairwise cosine similarity between each text embedding and each label embedding. For a given text embedding~$T_i$, we obtain a list of cosine similarity values~$C_i$ with length~$n$ that contains the similarity values between~$T_i$ and all label embeddings in~$L$. We also conduct experiments using Euclidean distance as an alternative similarity metric.

\[
\mathbf{C}_i = \left[ \text{cos}(\mathbf{T}_i, \mathbf{l}_1), \text{cos}(\mathbf{T}_i, \mathbf{l}_2), \dots, \text{cos}(\mathbf{T}_i, \mathbf{l}_n) \right]
\]

\begin{figure*}[t!]
\centering
\includegraphics[width = 0.75\textwidth]{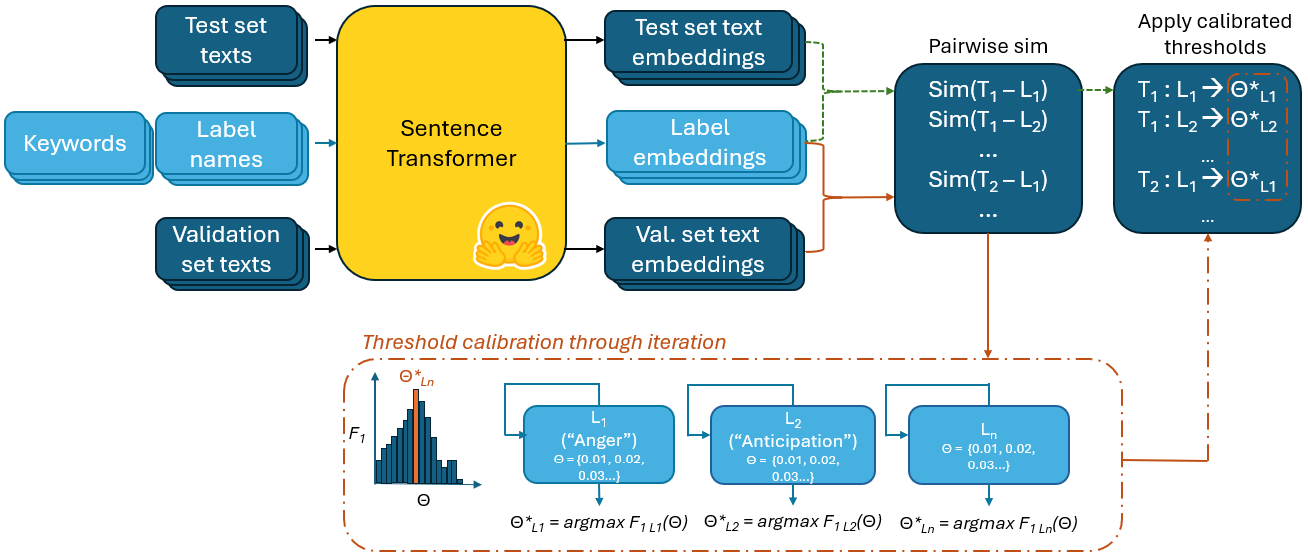}
\caption{Overview of the thresholding approach and Inference Stage. Texts and label representations are embedded, after which the cosine similarity is used to determine label relevance. Thresholds are optimized based on performance on a validation set.}
\label{fig:method}
\end{figure*}

As mentioned previously, a threshold~$\theta$ must be estimated to determine whether a label should be assigned to a text or not. If the similarity between~$T_i$ and the label embedding~$L_j$ equals or exceeds~$\theta$, the corresponding label is assigned to the text. We experiment with multiple thresholding mechanisms to determine~$\theta$:
\begin{enumerate}
    \item \textbf{0.5 and Normalized 0.5}: We experiment with two baseline thresholds. First, we use 0.5 as a threshold for all labels, regardless of the model or dataset. Second, we normalize the similarity distributions with min-max normalization as described in Section~\ref{sec:thresholding-methods} and apply 0.5 as a threshold.
    \item \textbf{Fine-tuned Uniform Thresholds}: With this method, the optimal threshold is determined by selecting the one that maximizes performance on the validation set. We iterate over thresholds from~$0.0$ to~$1.0$ in increments of~$0.01$, applying each threshold to assign labels. The macro-averaged~$F_1$-score is calculated for each threshold and the threshold with the highest score is chosen for final predictions on the test set.
    \item \textbf{Fine-tuned Label-Specific Thresholds}: With this method (cf. Figure~\ref{fig:method}), we approach the multi-label classification problem as a series of independent binary classification problems. We iterate over a range of thresholds between~$0.0$ and~$1.0$ with an increment of~$0.01$ and select the optimal threshold for each label separately based on the highest~$F_1$-score on the positive class for classifying the test set. For labels in the test set where no instance appears in the validation set, we assign the average of all fine-tuned thresholds.
    
\end{enumerate}

\subsection{Upper-bound baseline and Zero-shot with LLMs}
To compare our results with other existing methods, we fine-tune a \textit{RoBERTa} model~\cite{liu-etal-2019-roberta}\footnote{\url{https://huggingface.co/FacebookAI/roberta-base}} on the training splits of the datasets. Since the model is fine-tuned on the training data, we hypothesize that it will outperform the distance-based method. The learning rate is optimized by conducting grid-search experiments. Additionally, we prompt two LLMs, namely \textit{GPT-4o} and \textit{Gemini Pro 1.5}~\cite{geminiteam2024gemini15unlockingmultimodal}\footnote{Due to the limitations of the \textit{Gemini Pro 1.5} API in handling the size of nested objects during function calls, we were unable to process datasets with a larger number of labels, such as AAPD and Reuters.}, to perform zero-shot MLTC using the Binary Relevance method on the test data.

\subsection{Evaluation Metrics}
For the DBC experiments, we evaluate the performance of the different models and thresholding mechanisms using micro- and macro-averaged $F_1$~scores. In addition, we calculate the precision at~$k$ ($P@k$), where~$k=1$. This metric indicates whether the closest label to the text is a true label. 

\section{Results and Discussion}\label{sec:results}
\subsection{Exploratory Study}
The results of our experiments on analyzing the differences in similarity scales across embedding models (\textbf{H1}), the impact of domain-specific (inter-dataset) characteristics within a single model (\textbf{H2}), and the potential for class-specific similarity scales (\textbf{H3}) have partially confirmed our initial hypotheses, through the following findings. We address how each hypothesis is supported in the following sections.

\subsubsection{\textbf{Comparing Similarity Ranges Between Models (H1)}} \label{sec:hyp1}
Upon comparing the similarity ranges across various embedding models, we observed notable statistical differences in similarity distributions, challenging the suitability of a fixed universal threshold. Specifically, all models exhibit strong statistically significant differences ($p<.001$ for each possible model combination)  in distributions between each other. Additionally, median and mean similarity scores (cf. Table~\ref{tab:mean-median}\footnote{The results for all models can be found in Table~\ref{tab:mean-median-comp}, Appendix~\ref{app:hyp1}.}) vary considerably among models. For example, \textit{all-MPNet} stands out with a median score of 0.12, substantially staying under the commonly used threshold of 0.5, indicating a lower baseline similarity on the SemEval dataset. In contrast, other models like \textit{GTE-Large} and \textit{GIST-Large} exhibit much higher average similarity, with respective median scores of 0.48 and 0.35 on the same dataset. This variability underlines the need for model-specific threshold settings, as the applicability of a single threshold across diverse models can lead to inaccurate interpretations of semantic similarity.

Upon analyzing the similarity ranges, we considered applying min-max normalization to align the distributions across models. This method transforms scores to a common scale (0 to 1), possibly enabling a unified thresholding approach. While the min-max normalization has forced the similarity distribution to become more alike, the 0.5 threshold remains higher compared to the median of each distribution. For example, normalizing the scores of \textit{GTE-Large} and \textit{all-MPNet} brought their median values closer, making them more comparable.

These results can be attributed to differences in pre-training data and training methods used in the development of the embedding models. For example, \textit{GIST-Large} is trained on top of the \textit{BGE-Large} model using the MEDI dataset, according to the model's dataset card\footnote{\url{https://huggingface.co/avsolatorio/GIST-large-Embedding-v0}}. In contrast, \textit{Stella} uses distilled embeddings from larger models, such as \textit{Qwen}.

\begin{table*}[!ht]
\caption{(Normalized) Mean and Median Cosine Similarity per Model and Dataset}
\resizebox{\textwidth}{!}{%
\begin{tabular}{r|cccc|cccc|cccc|cccc|cccc}
\multicolumn{1}{l|}{} & 
\multicolumn{4}{c|}{\textbf{SemEval}} & 
\multicolumn{4}{c|}{\textbf{BioTech}} & 
\multicolumn{4}{c|}{\textbf{Reuters}} & 
\multicolumn{4}{c|}{\textbf{AAPD}} & 
\multicolumn{4}{c}{\textbf{LitCOVID}} \\ \hline

\multicolumn{1}{r|}{model} & 
\multicolumn{1}{l}{mn} & \multicolumn{1}{l}{\begin{tabular}[c]{@{}c@{}}mn \\ (norm)\end{tabular}} & \multicolumn{1}{l}{mdn} & \multicolumn{1}{l|}{\begin{tabular}[c]{@{}c@{}}mdn \\ (norm)\end{tabular}} & 
\multicolumn{1}{l}{mn} & \multicolumn{1}{l}{\begin{tabular}[c]{@{}c@{}}mn \\ (norm)\end{tabular}} & \multicolumn{1}{l}{mdn} & \multicolumn{1}{l|}{\begin{tabular}[c]{@{}c@{}}mdn \\ (norm)\end{tabular}} & 
\multicolumn{1}{l}{mn} & \multicolumn{1}{l}{\begin{tabular}[c]{@{}c@{}}mn \\ (norm)\end{tabular}} & \multicolumn{1}{l}{mdn} & \multicolumn{1}{l|}{\begin{tabular}[c]{@{}c@{}}mdn \\ (norm)\end{tabular}} & 
\multicolumn{1}{l}{mn} & \multicolumn{1}{l}{\begin{tabular}[c]{@{}c@{}}mn \\ (norm)\end{tabular}} & \multicolumn{1}{l}{mdn} & \multicolumn{1}{l|}{\begin{tabular}[c]{@{}c@{}}mdn \\ (norm)\end{tabular}} & 
\multicolumn{1}{l}{mn} & \multicolumn{1}{l}{\begin{tabular}[c]{@{}c@{}}mn \\ (norm)\end{tabular}} & \multicolumn{1}{l}{mdn} & \multicolumn{1}{l}{\begin{tabular}[c]{@{}c@{}}mdn \\ (norm)\end{tabular}} \\ \hline

GIST-Large & 0.36 & 0.31 & 0.35 & 0.30 & 0.32 & 0.38 & 0.32 & 0.38 & 0.35 & 0.29 & 0.34 & 0.37 &  0.34 & 0.37 & 0.33 & 0.33 & 0.32 & 0.33 & 0.31 & 0.29 \\
GTE-Large & 0.48 & 0.44 & 0.48 & 0.44 & 0.45 & 0.42 & 0.45 & 0.43 & 0.36 & 0.43 & 0.35 & 0.46 & 0.43 & 0.48 & 0.43 & 0.41 & 0.44 & 0.41 & 0.43 & 0.44 \\
Stella & 0.35 & 0.25 & 0.34 & 0.24 & 0.28 & 0.24 & 0.28 & 0.23 & 0.26 & 0.26 & 0.25 & 0.25 & 0.33 & 0.35 & 0.32 & 0.35 & 0.33 & 0.34 & 0.31 & 0.27 \\
all-MPNet & 0.13 & 0.30 & 0.12 & 0.29 & 0.14 & 0.36 & 0.13 & 0.35 & 0.29 & 0.07 & 0.27 & 0.22 & 0.31 & 0.10 & 0.30 & 0.12 & 0.33 & 0.11 & 0.32 & 0.09 \\
GloVe & 0.23 & 0.39 & 0.22 & 0.38 & 0.45 & 0.60 & 0.46 & 0.60 & 0.45 & 0.19 & 0.44 & 0.47 & 0.60 & 0.47 & 0.61 & 0.48 & 0.65 & 0.48 & 0.65 & 0.21 \\

\end{tabular}%
}
\label{tab:mean-median}
\end{table*}

\subsubsection{\textbf{Domain-Specific Variability (H2)}}
\begin{figure}[t!]
\centering
\includegraphics[width=0.8\linewidth]{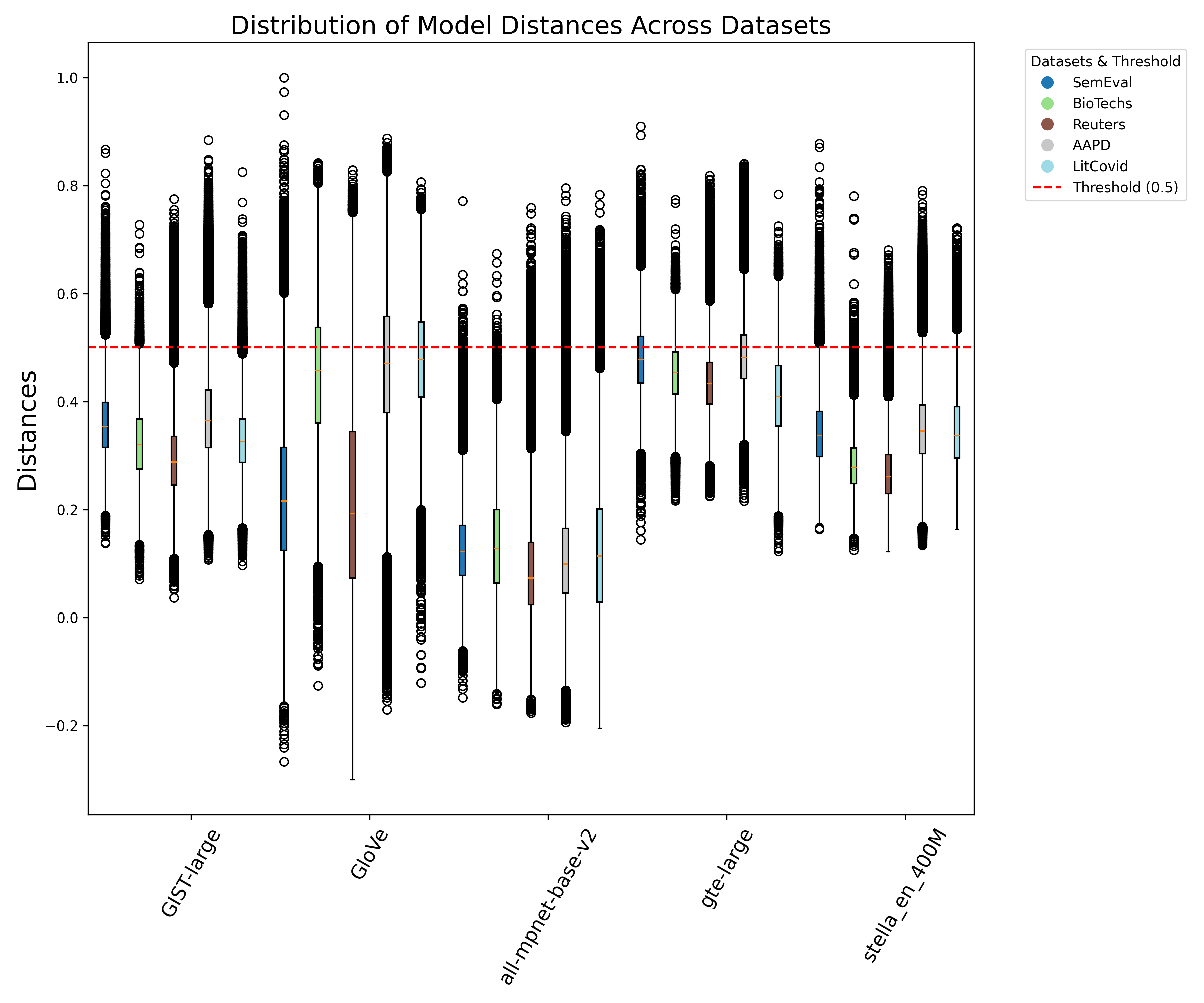}
\caption{Bar charts showing the variation in (non-normalized) cosine similarity distributions between individual models and domains.}
\label{fig:domain_variability}
\end{figure}

The experimental findings also support the hypothesis of domain-specific variability affecting semantic similarity scales within a single model. We find that the differences in distribution across all datasets are statistically significant, except for \textit{Stella}. For this model, only the similarity distribution between SemEval and LitCovid did not differ significantly. Concerning the variability, for instance, the \textit{GTE-Large} model exhibited a median similarity score of 0.48 for the SemEval dataset, while for the Reuters dataset, the median score was lower at 0.35. This variation across domains suggests that even within a single model, the perceived similarity between texts can significantly change depending on the domain. Figure~\ref{fig:domain_variability} illustrates this domain variability by plotting the cosine similarity between each text in a dataset and all labels. After normalizing scores within each domain, we observed that differences tend to diminish (cf. Appendix~\ref{app:hyp1}, Figure~\ref{fig:norm_domain_variability}). For example, normalization could raise the lower median similarity score for the Reuters dataset, making it more comparable to scores from other domains.

The variability across domains can be attributed to differences in the underlying text distributions of pre-training or post-training data. Some strands of text data (domains) may be underrepresented in terms of word frequency compared to others. Since word frequency plays a role in the robustness of model representations~\cite{zhou2022problemscosinemeasureembedding}, this can lead to differences in similarity scores.

\subsubsection{\textbf{Class-specific Similarity Scales (H3)}}
When comparing the similarity scales across different classes within a single model, we observed statistically significant differences in similarity distributions in most cases across all models and almost all datasets, except for the Reuters dataset (cf. Table~\ref{tab:results-class-dists}). This could be explained by the relatively large and detailed label set including closely related labels, causing some labels to be clustered together in the embedding space and resulting in overlapping cosine similarity distributions. 

Similar to the results for \textbf{H2}, these findings may be explained by the effect of word frequency on how certain words are encoded in a model: infrequent words have weaker representations than common ones, thereby influencing the similarity scores observed in our experiments. 

As a result, the experimental findings for \textbf{H1–H3} question the suitability of a fixed universal threshold.

\begin{table}[t!]
\caption{Proportion of statistically significant ($p<.05$) differences between cosine similarity distributions of label pairs in a dataset, per model.}
\centering
\resizebox{0.60\columnwidth}{!}{%
\begin{tabular}{r|ccccc}
\textbf{Model} & 
\textbf{SemEval} & 
\textbf{BioTech} & 
\textbf{Reuters} & 
\textbf{AAPD} & 
\textbf{LitCovid} \\ \hline
GIST-Large & 0.78 & 0.40 & 0.07 & 0.89 & 0.81 \\
GTE-Large & 0.85 & 0.35 & 0.09 & 0.87 & 0.95 \\
Stella & 0.75 & 0.50 & 0.10 & 0.90 & 1.00 \\
UAE-Large & 0.82 & 0.35 & 0.07 & 0.90 & 0.95 \\
Mxbai & 0.80 & 0.38 & 0.07 & 0.89 & 0.95 \\
BGE & 0.82 & 0.34 & 0.09 & 0.92 & 1.00 \\
all-MPNet & 0.71 & 0.46 & 0.10 & 0.90 & 1.00 \\
all-MPNet-Wiki & 0.67 & 0.42 & 0.08 & 0.88 & 0.95 \\
GloVe & 0.98 & 0.67 & 0.28 & 0.93 & 0.95
\end{tabular}%
}
\label{tab:results-class-dists}
\end{table}

\subsection{Distance-based MLTC}

\subsubsection{\textbf{Comparing Thresholding Approaches (H4)}}\label{sec:thresholding-methods} \hfill\
 

\textbf{\textit{General Results.}}
Overall, we observe that optimizing thresholds for each label separately consistently yields considerable improvements compared to using a uniform threshold --either the commonly used 0.5 and an optimized-- across all evaluated sentence encoders (cf.~Table~\ref{tab:classification-results}\footnote{Results for all sentence encoders and for Euclidean distance are provided in Tables~\ref{tab:all-classification-results} and~\ref{tab:all-results-euclid} in Appendix~\ref{app:comp-results}.}). This finding aligns with our exploratory study, indicating that the similarity ranges for labels are unique and should be considered when performing distance-based classification. Optimizing each label threshold helps to enhance classification by capturing the specific characteristics of each label, leading to better overall performance with average improvements of 52\% and 14\% in macro-$F_1$ over the 0.5 and optimized uniform threshold, respectively. Regarding the baselines, fine-tuned models show superior performance, followed by zero-shot models, with fine-tuning label-specific thresholds in some cases matching or even exceeding the performance of fine-tuned models.

\textbf{\textit{Comparison with Baselines.}}
The results of the classification experiments are summarized in Table~\ref{tab:classification-results-baselines}, which contains the micro-$F_1$ (mi$F_1$), macro-$F_1$ (ma$F_1$), and Precision@1$P@1$ scores for each model and thresholding approach. When comparing distance-based methods (with optimized thresholds for each label) to zero-shot and fine-tuned model baselines, we observe several patterns. When trained on the full training split, \textit{RoBERTa} consistently outperforms all zero-shot and distance-based methods (with optimized thresholds) in terms of ma$F_1$ and mi$F_1$ across the AAPD, and LitCovid datasets. For mi$F_1$, \textit{RoBERTa} also surpasses them on SemEval and Reuters datasets. However, when RoBERTa is trained on a sample, it is generally outperformed by the distance-based method with optimized thresholds in terms of ma$F_1$, except for the LitCovid dataset. For mi$F_1$, this advantage is observed only in the AAPD dataset.

In zero-shot setup, \textit{GPT-4o} achieves the highest ma$F_1$ on the BioTech and Reuters datasets, while \textit{Gemini Pro 1.5} performs best on the SemEval dataset in terms of ma$F_1$. Notably, the distance-based method with optimized thresholds outperforms Gemini Pro 1.5 in mi$F_1$ on the BioTech and LitCovid datasets, and also surpasses \textit{GPT-4o} in mi$F_1$ on the AAPD dataset.

\textbf{\textit{Label-specific vs. Uniform Thresholds.}} When comparing the uniform fine-tuned threshold with label-specific ones, we observe the largest increases in ma$F_1$ and mi$F_1$ on the BioTech and LitCovid datasets. SemEval, Reuters, and AAPD also demonstrate considerable improvements, though not as substantial as those observed on the previously mentioned datasets. This also highlights that the dynamics between each label and text are unique, which should be taken into account when performing distance-based classification. These interactions are not considered in uniform thresholding methods, as explored in~\cite{sarkaretal2023zero}. An exception is observed for \textit{GIST-Large}, which does not benefit from label-specific thresholds on the Reuters dataset. This could be explained by the absence of some labels from the test set in the validation set. Since we estimated the optimal thresholds for these labels by averaging all optimized thresholds, they were therefore not optimally calculated in this case. The $P@1$ scores are the same for both thresholding methods. 

\begin{figure*}[t!]
    \centering
    \begin{minipage}{0.46\textwidth}
        \centering
        \includegraphics[width=\linewidth]{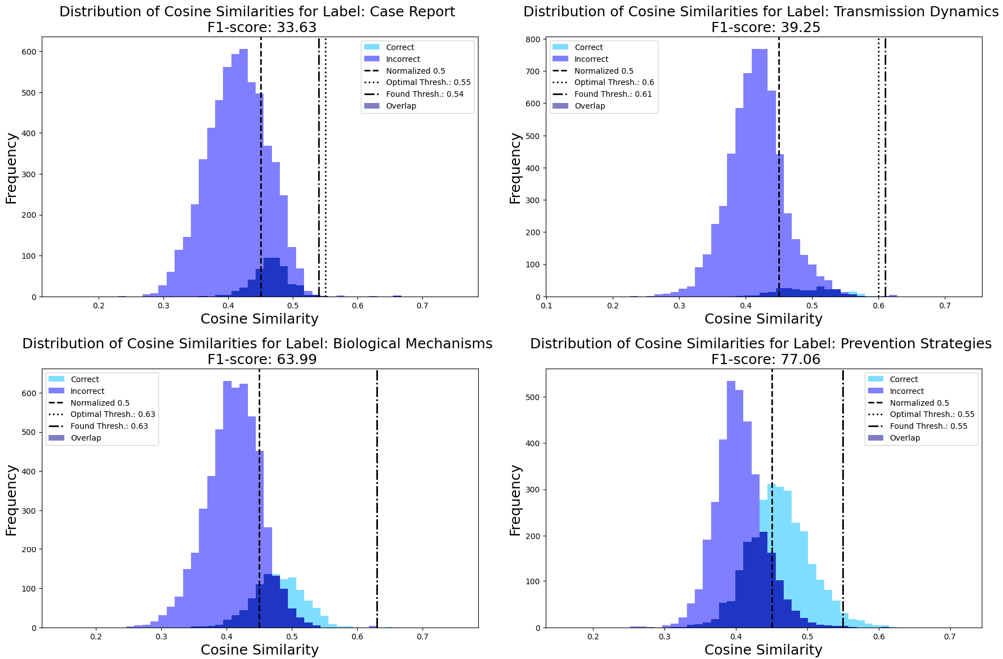}
        \caption{Distributions of cosine similarity scores per label of the LitCovid dataset, obtained using \textit{Stella}. }
        \label{fig:dists-litcovid}
    \end{minipage}%
    \hspace{0.04\textwidth}
    \begin{minipage}{0.445\textwidth}
        \centering
        \includegraphics[width=\linewidth]{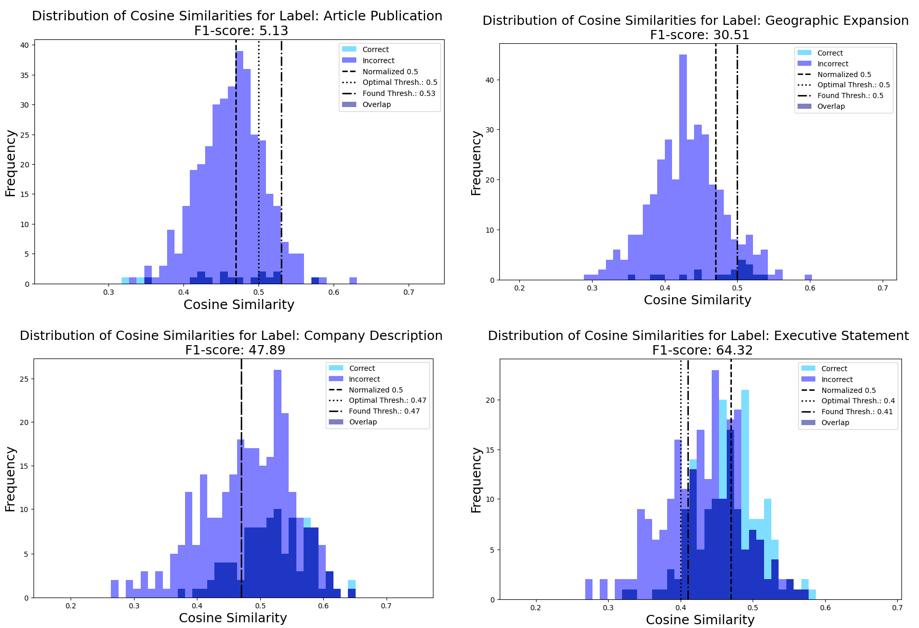}
        \caption{Distributions of cosine similarity Scores per label of the BioTech dataset, obtained using \textit{GIST-Large}. }
        \label{fig:dists-biotech}

    \end{minipage}
\end{figure*}

In general, our proposed thresholding approach approximates the optimal thresholds, as evidenced in Figure~\ref{fig:dists-litcovid}. The optimal thresholds are calculated by performing a similar threshold optimization process on the test sets, instead of the validation sets, as described in Section~\ref{sec:dbc}. For example, the optimized thresholds for ``Case Report'', ``Biological Mechanisms'', and ``Prevention Strategies'' are close to or equal to the optimal thresholds.

\textbf{\textit{Best-performing Embedding Models.}}
Overall, we observe that \textit{Stella} yields the best performance on the Reuters, AAPD, and LitCovid datasets. On the other datasets, \textit{GIST-Large} (SemEval) and \textit{GTE-Large} (BioTech) achieve the best performance. The reason that \textit{GTE-Large} performs best on BioTech can be attributed to its longer context length, as this dataset contains texts that, on average, are longer than 512 tokens (cf.~Table~\ref{tab:datasets}). The additional information encoded at the ends of the texts therefore still proves useful for classification. Although the performance of non-contextual embedding models such as \textit{GloVe} remains respectable, they are consistently among the worst-performing models (cf. Appendix~\ref{app:comp-results}, Table~\ref{tab:all-classification-results}).

\begin{table*}[!ht]
\centering
\caption{Classification Results of Different Thresholding Methods on All Datasets. An Asterisk (*) Indicates Best Performance with Label Names.}
\label{tab:classification-results}
\resizebox{1\textwidth}{!}{%
\begin{tabular}{c|r|rrr|rrr|rrr|rrr|rrr|c}
\multicolumn{1}{l}{}& 
\multicolumn{1}{c|}{} & 
\multicolumn{3}{c|}{\textbf{SemEval}} & 
\multicolumn{3}{c|}{\textbf{BioTech}} & 
\multicolumn{3}{c|}{\textbf{Reuters}} & 
\multicolumn{3}{c|}{\textbf{AAPD}} & 
\multicolumn{3}{c}{\textbf{LitCovid}} & 
\multicolumn{1}{c}{\textbf{Avg}} \\

\hline
\multicolumn{1}{r|}{\textbf{model}} & 
\multicolumn{1}{l|}{\textbf{thr}} & 
\multicolumn{1}{l}{\textbf{ma$F_1$}} & \multicolumn{1}{l}{\textbf{mi$F_1$}}  & \multicolumn{1}{l|}{\textbf{$P@1$}} & 
\multicolumn{1}{l}{\textbf{ma$F_1$}} & \multicolumn{1}{l}{\textbf{mi$F_1$}}& \multicolumn{1}{l|}{\textbf{$P@1$}} & 
\multicolumn{1}{l}{\textbf{ma$F_1$}} & \multicolumn{1}{l}{\textbf{mi$F_1$}} & \multicolumn{1}{l|}{\textbf{$P@1$}} & 
\multicolumn{1}{l}{\textbf{ma$F_1$}} & \multicolumn{1}{l}{\textbf{mi$F_1$}} & \multicolumn{1}{l|}{\textbf{$P@1$}} & 
\multicolumn{1}{l}{\textbf{ma$F_1$}} & \multicolumn{1}{l}{\textbf{mi$F_1$}}& \multicolumn{1}{l|}{\textbf{$P@1$}} &
\multicolumn{1}{c}{\textbf{avg ma$F_1$}} \\ 

\hline
\multirow{4}{*}{\textbf{GIST-Large}} & 0.5 & 42.2  & 48.58& 67.51 & 18.6& 19.3  & 28.1  & 32.8  & 37.92  & 50.55  & 28.46 & 30.02  & 64.7  & 35.35 & 42.95 & 61.07 & 31.48 \\
  & n0.5& 42.94 & 49.53& 67.51 & 15.43  & 19.0  & 28.1  & 12.3  & 15.81  & 50.55  & 30.89 & 32.64  & 64.7  & 40.76 & 48.47 & 61.07 & 28.46  \\
  & unif& 44.99 & 51.39& 67.51 & 18.54  & 21.5  & 28.1  & \textbf{32.8}  & 37.92  & 50.55 & 38.21 & 43.11  & 64.7  & 40.32 & 47.41 & 61.07 & 34.97  \\
  & lbl & {\cellcolor[rgb]{0.937,0.937,0.937}}\textbf{47.59} & {\cellcolor[rgb]{0.937,0.937,0.937}}\textbf{55.0} & {\cellcolor[rgb]{0.937,0.937,0.937}}\textbf{67.51} & \textbf{23.29}  & \textbf{33.0}  & \textbf{28.1}  & 31.6* & \textbf{38.02*} & \textbf{50.55*} & \textbf{41.06*}& {\cellcolor[rgb]{0.937,0.937,0.937}}\textbf{49.97*} & {\cellcolor[rgb]{0.937,0.937,0.937}}\textbf{64.3*} & \textbf{51.34*}& \textbf{56.31*}& \textbf{44.96*} & \textbf{38.98} \\ 
\hline
\multirow{4}{*}{\textbf{GTE-Large}}  & 0.5 & 37.25 & 39.42& 61.95 & 13.95  & 16.7  & 26.8  & 8.3& 11.44  & 48.76 & 17.51*& 18.91* & 61.0* & 38.82 & 44.07 & 60.65 & 23.17 \\
  & n0.5& 41.23 & 45.84& 61.95 & 13.32  & 15.9  & 26.8  & 8.86  & 12.2& 48.76 & 23.98 & 27.02  & 62.0  & 39.76 & 47.55 & 60.65 & 25.43 \\
  & unif& 41.35 & 46.79& 61.95 & 19.56  & 21.0  & 26.8  & 33.2* & 38.13* & 48.76* & 33.88 & 39.49  & 62.0  & 43.64*& 48.53*& 51.16* & 34.33 \\
  & lbl & \textbf{44.01} & \textbf{50.87}& \textbf{61.95} & {\cellcolor[rgb]{0.937,0.937,0.937}}\textbf{26.18*} & {\cellcolor[rgb]{0.937,0.937,0.937}}\textbf{36.6*} & {\cellcolor[rgb]{0.937,0.937,0.937}}\textbf{26.8*} & \textbf{34.5*} & \textbf{41.31*} & \textbf{48.76*} & \textbf{37.41} & \textbf{40.13}  & \textbf{62.0}  & \textbf{48.79*}& \textbf{55.71*}& \textbf{51.16*} & \textbf{38.18} \\ 
\hline
\multirow{4}{*}{\textbf{Stella}}& 0.5 & 28.91 & 32.37& 61.83 & 6.29& 2.34  & 28.9  & 29.2  & 29.41  & 69.04  & 35.19 & 43.68  & 64.1  & 29.2  & 28.36 & 55.47 & 25.76 \\
  & n0.5& 25.62 & 28.45& 61.83 & 18.03  & 19.9  & 28.9  & 20.9  & 29.83  & 69.04  & 31.81*& 35.07* & 56.8* & 43.68 & 46.43 & 55.47 & 28.01 \\
  & unif& 41.54 & 45.9 & 61.83 & 18.25  & 19.4  & 28.9  & 36.5* & 54.26* & 69.04* & 35.19 & 43.68  & 64.1  & 44.5  & 47.4  & 55.47 & 35.20 \\
  & lbl & \textbf{42.54} & \textbf{48.66}& \textbf{61.83} & \textbf{26.02}  & \textbf{35.3}  & \textbf{28.9}  & {\cellcolor[rgb]{0.937,0.937,0.937}}\textbf{37.1*} & {\cellcolor[rgb]{0.937,0.937,0.937}}\textbf{55.54*} & {\cellcolor[rgb]{0.937,0.937,0.937}}\textbf{69.04*} & {\cellcolor[rgb]{0.937,0.937,0.937}}\textbf{42.41} & \textbf{48.16}  & \textbf{64.1}  & {\cellcolor[rgb]{0.937,0.937,0.937}}\textbf{57.71} & {\cellcolor[rgb]{0.937,0.937,0.937}}\textbf{63.88} & {\cellcolor[rgb]{0.937,0.937,0.937}}\textbf{55.47} & \textbf{41.16} \\ 
\hline
\multirow{4}{*}{\textbf{UAE-Large}}& 0.5& 42.08*& 48.1*& 64.34*& 16.85  & 20.7  & 25.2  & 22.7  & 27.39  & 46.44  & 12.85*& 14.09* & 62.5* & 31.44 & 33.96 & 56.61 & 25.18 \\
  & n0.5& 40.98 & 45.65& 63.82 & 15.15  & 18.4  & 25.2  & 12.1  & 16.0& 46.44  & 21.78 & 22.68  & 64.0  & 35.58 & 41.7  & 56.61 & 25.12 \\
  & unif& 42.1  & 48.69& 63.82 & 17.95  & 20.8  & 25.2  & 31.4* & 32.4*  & 46.44* & 33.28 & 38.15  & 64.0  & 35.36 & 40.97 & 56.61 & 32.02 \\
  & lbl& \textbf{44.05*}& \textbf{53.07*}  & \textbf{64.34*}& \textbf{22.55}  & \textbf{32.49} & \textbf{25.2}  & \textbf{34.0*} & \textbf{35.37*} & \textbf{46.44*} & \textbf{38.6}  & \textbf{43.86}  & \textbf{64.0}  & \textbf{45.77*}& \textbf{53.89*}& \textbf{39.77*} & \textbf{36.99} \\ 
\hline
\multirow{4}{*}{\textbf{BGE}}& 0.5& 39.51*& 42.56*  & 57.32*& 15.57* & 17.18*& 17.85*& 11.4* & 13.34  & 51.04  & 11.03*& 11.69* & 49.0* & 34.94*& 38.4* & 53.55* & 22.49\\
  & n0.5  & 40.35*& 43.83*  & 57.32*& 15.13* & 17.3* & 17.85*& 8.36  & 9.95& 51.04  & 23.39 & 24.09  & 49.6  & 35.4* & 39.3* & 53.55* & 24.53 \\
  & unif& 39.31 & 44.93& 61.64 & 16.77  & 19.5  & 21.5  & 32.1* & 35.82* & 51.04* & 31.62 & 36.64  & 61.4  & 40.53*& 45.24*& 53.55* & 32.07 \\
  & lbl & \textbf{42.68} & \textbf{48.97}& \textbf{61.64} & \textbf{21.63}  & \textbf{30.4}  & \textbf{21.5}  & \textbf{33.0*} & \textbf{43.43*} & \textbf{51.04*} & \textbf{37.05} & \textbf{43.97}  & \textbf{61.4}  & \textbf{48.08*}& \textbf{56.53*}& \textbf{53.55*} & \textbf{36.49} \\ 
\end{tabular}%
}
\end{table*}

\begin{table*}
\centering
\caption{Classification Results from Fine-tuned models and Generative LLMs on all Datasets, Compared with Best-Performing Distance-Based Classification (DBC) Model.}
\label{tab:classification-results-baselines}
\resizebox{0.9\textwidth}{!}{%
\begin{tabular}{l|rr|rr|rr|rr|rr|cc} & 
\multicolumn{2}{c|}{\textbf{SemEval}} & 
\multicolumn{2}{c|}{\textbf{BioTech}} &
\multicolumn{2}{c|}{\textbf{Reuters}} & 
\multicolumn{2}{c|}{\textbf{AAPD}} & 
\multicolumn{2}{c|}{\textbf{LitCovid}} &
\multicolumn{2}{c}{\textbf{Avg}} \\
\hline
\multicolumn{1}{c|}{\textbf{approach/model}} & 
\multicolumn{1}{l}{\textbf{ma$F_1$}} & \multicolumn{1}{l|}{\textbf{mi$F_1$}} & 
\multicolumn{1}{l}{\textbf{ma$F_1$}} & \multicolumn{1}{l|}{\textbf{mi$F_1$}} & 
\multicolumn{1}{l}{\textbf{ma$F_1$}} & \multicolumn{1}{l|}{\textbf{mi$F_1$}} & 
\multicolumn{1}{l}{\textbf{ma$F_1$}} & \multicolumn{1}{l|}{\textbf{mi$F_1$}} & 
\multicolumn{1}{l}{\textbf{ma$F_1$}} & \multicolumn{1}{l|}{\textbf{mi$F_1$}} &
\multicolumn{1}{c}{\textbf{avg ma$F_1$}} & \multicolumn{1}{c}{\textbf{avg mi$F_1$}}\\ 
\hline
\textbf{DBC} & 47.59 & 55.0 & 26.18 & 36.6 & 37.1 & 55.54 & 41.06 & 49.97 & 57.71 & 63.88 & 41.93 & 52.20 \\
\textbf{Roberta (sample)} & 44.22 & 63.83 & 16.64 & \textbf{47.92} & 19.2 & 83.23 & 11.77 & 47.99 & 72.45 & 86.10 & 32.86 & 65.81 \\
\textbf{Roberta} & 51.83 & \textbf{69.40} & 13.23 & 47.46 & 38.7 & \textbf{89.12} & \textbf{52.99} & \textbf{72.02} & \textbf{81.93} & \textbf{88.28} & 47.74 & \textbf{73.26} \\
\textbf{GPT-4o} & 52.72 & 60.99 & \textbf{36.84} & 43.4 & \textbf{59.5} & 76.01 & 42.72 & 45.62 & 63.6 & 66.35 & \textbf{51.08} & 58.47 \\
\textbf{Gemini Pro 1.5} & \textbf{54.73} & 59.24 & 33.9 & 34.5 & N/A & N/A & N/A & N/A & 49.72 & 56.46 & N/A & N/A
\end{tabular}%
}
\end{table*}

\begin{figure}[t]
\centering
\includegraphics[width =0.7\columnwidth]{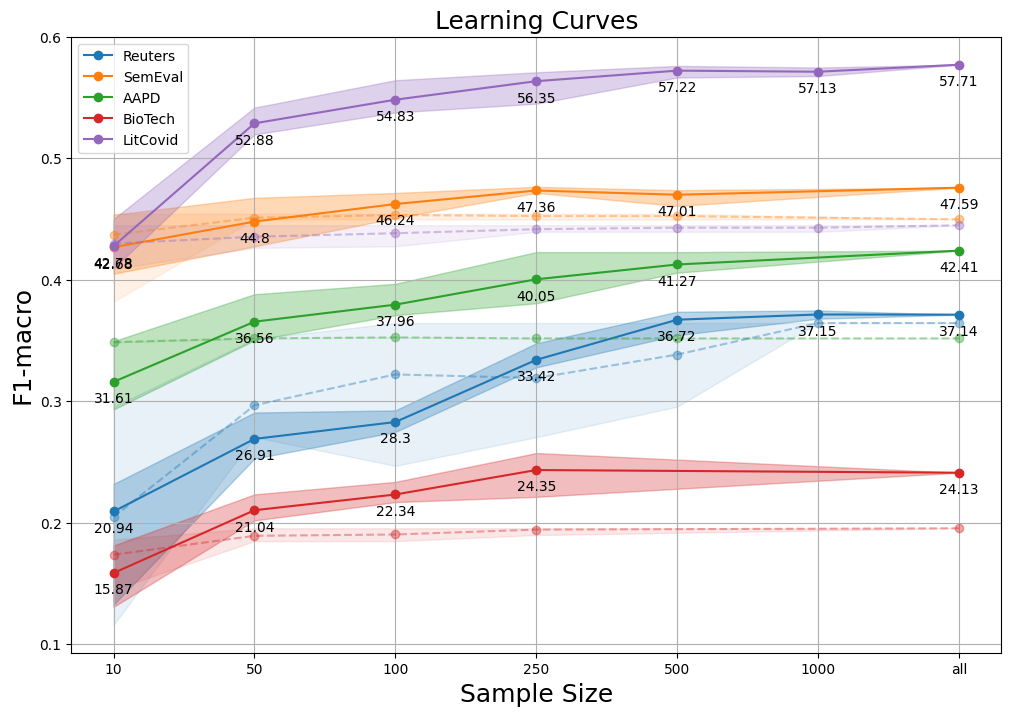}
\caption{Results from learning curve experiments with the best performing model on each dataset (\textit{GIST-Large} for SemEval, \textit{GTE-Large} for BioTech and \textit{Stella} for all other datasets). The dashed lines show the results conducted with uniform fine-tuned thresholds. Five random samples for each size are taken.}
\label{fig:learning-curve}
\end{figure}

\textbf{\textit{Effect of Sample Size.}}
Additionally, we conducted learning curve experiments by taking five random stratified splits \cite{mlstratification} of the validation data for each sample size to evaluate how much data is needed to obtain accurate thresholds. We calculate label-specific thresholds as before, but on smaller samples of the validation data. The results --as summarised in Figure~\ref{fig:learning-curve}-- suggest that more data leads to more accurate thresholds and, therefore, better performance. Notably, for all datasets except Reuters, label-specific thresholds derived from just 50-100 validation samples achieve equal or superior ma$F_1$ scores compared to an optimized uniform threshold derived from the entire validation dataset. Using only 10 examples per label achieves 74\% of the full method's performance, while 100 examples reach 91\% of full performance.  These results suggest that our proposed method is more attractive for scenarios where limited annotated data is available.

\subsubsection{\textbf{Effect of Label Representation Methods}}
In general, we observe that enhancing label representations either by adjusting the label names or generating additional keywords improves classification results. Adjusting label names aids the separation between labels in the embedding space, therefore leading to better performance on all datasets. Tables~\ref{tab:lbl-name-comp-datasets} and~\ref{tab:lbl-name-comp-litcovid} in Appendix~\ref{app:label-adj} show the differences in performance.

The averaged keyword embedding as a label representation often yields improvements, suggesting that the centroid of the label and its corresponding keywords provides a more robust representation of the label's meaning. The most notable improvements are observed in terms of $P@1$, regardless of whether keywords yield better $F_1$-scores for a specific model. For a complete overview of results regarding label representations, consult Table~\ref{tab:results-label-rep} in Appendix~\ref{app:label-adj}. We expected that the relative score increase between label names and keyword embeddings would be the highest for word embedding models~\cite{Kosar-etal-2022}, but we did not observe any notable difference compared to other, more advanced sentence transformers. However, \textit{GloVe} benefits more consistently from keyword embeddings than other sentence transformers.

An exception to this is the Reuters dataset, where no model benefits from generated keywords, likely due to the large label space of the dataset. Since all labels belong to the same domain, the keywords generated by an LLM are likely to introduce semantic overlap between labels on the one hand, or be too generic for the label name, thereby introducing noise\footnote{We experimented with filtering the keywords further using LLMs, though this did not yield any improvements.}.

While these findings generally indicate that keyword-based representations can enhance performance, the effect is not uniform across datasets or models. For example, \textit{GIST-Large} benefits from averaged keyword embeddings on SemEval and BioTech, but not on AAPD or LitCovid. Similarly, \textit{GTE-Large} yields better results with averaged keyword representations on all datasets except BioTech and LitCovid. This could be explained by the fact that the quality of the label names is already satisfactory for those models to perform classification, and the keywords may introduce noise.

\subsubsection{\textbf{Error Analysis}}\label{sec:distributions}\hfill\\

\indent\textbf{\textit{Similarity Distributions.}} 
To further analyze the results of the experiments and the effectiveness of our proposed thresholding approach, we plot the cosine similarities between all text embeddings from the test data and the label embeddings. Given label $i$, the similarities are divided into two groups: similarities between label $i$'s embedding and texts where $i$ is a true label ($\alpha$), and similarities with texts where $i$ is not a true label ($\beta$). By plotting the similarities this way (cf. Figures~\ref{fig:dists-litcovid} and~\ref{fig:dists-biotech}), we can visualize the overlap between similarity distributions and, consequently, assess the effectiveness and feasibility of applying label-specific thresholds. For LitCovid (Figure~\ref{fig:dists-litcovid}), it can be observed that a clearer separation between $\alpha$ and $\beta$ leads to a higher $F_1$-score for a given label. The Pearson correlation coefficient between the overlap per class --as shown in Figure~\ref{fig:dists-litcovid}-- and the $F_1$-scores per class indicates that there is a very strong negative correlation (-0.97). We also find similar strong correlations for SemEval (-0.89) and AAPD (-0.71), in addition to weaker negative correlations for Reuters (-0.44) and BioTech (-0.55).

While distance-based classification works well on the LitCovid dataset, this classification method is more challenging on BioTech, as evidenced by the low performance scores from each model (cf.~Table~\ref{tab:classification-results}). It should also be noted that this dataset is challenging for LLMs and fine-tuned transformers as well. The challenge for distance-based methods stems from the completely overlapping distributions (Figure~\ref{fig:dists-biotech}), rendering it virtually impossible to select a threshold that can reliably separate texts where the label is true from those where it is false. However, the number of instances per class also affects the performance on the BioTech dataset: the more frequently a label occurs, the higher the $F_1$-score. This is further supported by a high Pearson correlation coefficient of 0.77.

\begin{figure*}[h!]
\centering
\includegraphics[width = 0.85\textwidth]{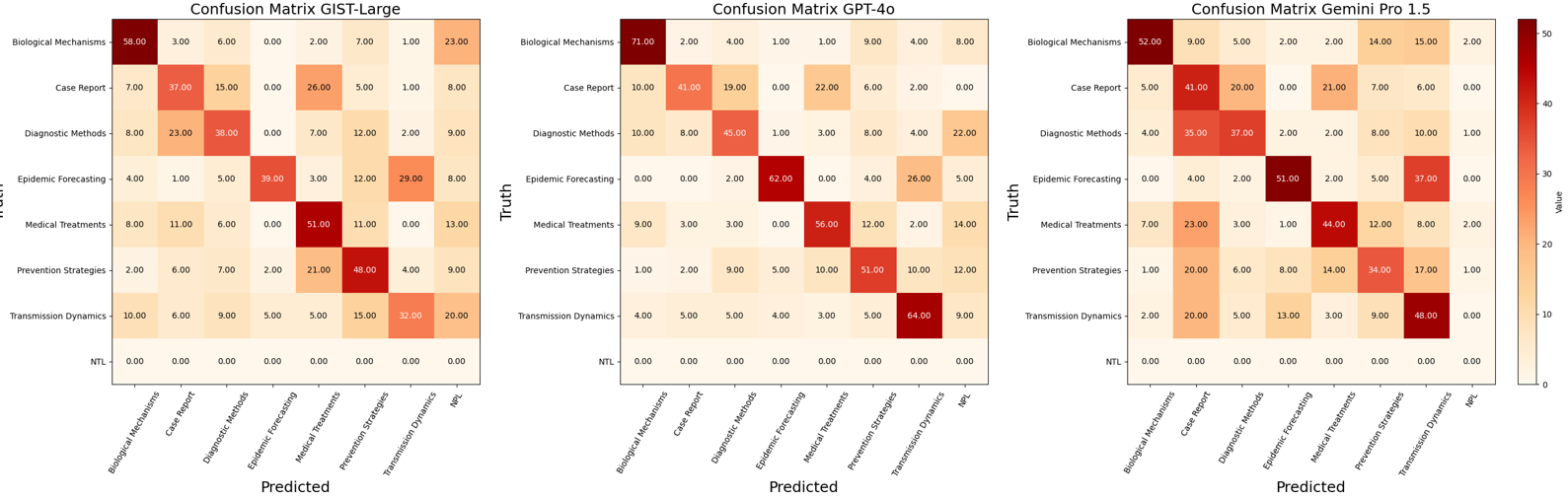}
\centering
\caption{Confusion Matrices of \textit{Stella}, \textit{GPT-4o} and \textit{Gemini Pro 1.5} on the LitCovid Dataset.}
\label{fig:cm-litcovid}
\end{figure*} 

\begin{figure*}[h!]
\centering
\includegraphics[width = 0.85\textwidth]{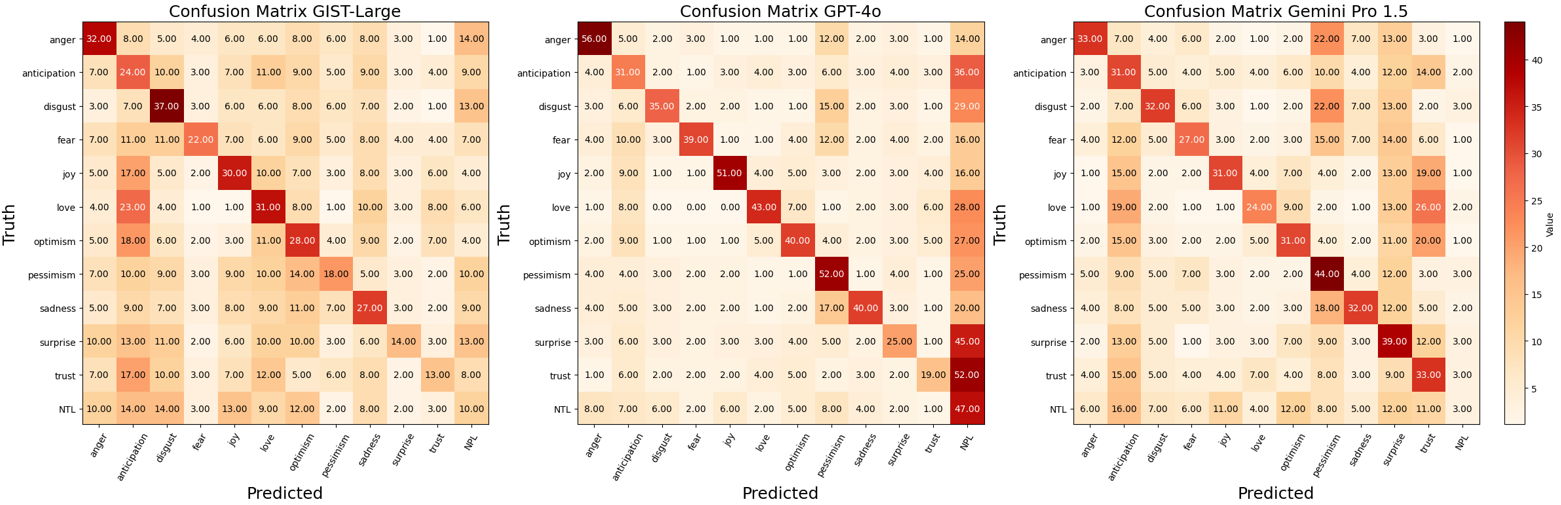}
\centering
\caption{Confusion Matrices of \textit{Stella}, \textit{GPT-4o} and \textit{Gemini Pro 1.5} on the SemEval Dataset.}
\label{fig:cm-semeval}
\end{figure*} 

\textbf{\textit{Confusion Matrices.}} We visualize the confusion matrices obtained from the best-performing sentence encoder, \textit{GPT-4o}, and \textit{Gemini Pro 1.5} using the method proposed by~\cite{mlcm}. In contrast to confusion matrices from binary or multi-class classification problems, an extra row and column are added, respectively indicating cases where no true labels are present or no labels are predicted. For LitCovid, we observe that the distance-based method makes similar mistakes to \textit{GPT-4o}. For example, both models frequently confuse ``Case Report''  with ``Medical Treatments'', ``Epidemic Forecasting'' with ``Transmission Dynamics'', and ``Preventive Strategies'' with ``Medical Treatments''. As evidenced by the lower performance of \textit{Gemini Pro 1.5} shown in Table~\ref{tab:classification-results}, the model confuses more labels with each other, which is particularly apparent for ``Case Report''. This label appears to serve as a fallback during the model’s prediction process.

On the SemEval dataset, we observe that the distance-based method underperforms compared to LLMs. One reason, according to the confusion matrix in Figure~\ref{fig:cm-semeval} is that the model often confuses the label ``anticipation'' with other labels. It could be hypothesized that this label is close to other labels in the embedding space due to the abstract nature of the emotion, especially compared to other emotions like ``anger''. To verify this, we embedded the emotion labels (and corresponding keywords) using \textit{GIST-Large} and calculated the average cosine similarity to all other labels in the dataset (cf.~Table~\ref{tab:sim-other-labels}). We found that, indeed, ``anticipation'' is more similar to other labels, albeit by a slight margin. This could cause confusion during classification, thereby introducing a high number of false positives. Interestingly, \textit{Gemini Pro 1.5} also experiences difficulties with the same label, in addition to ``trust'', ``surprise'', and ``pessimism''. \textit{GPT-4o}, on the other hand, excels at predicting correct subsets of true labels, as evidenced by the highlighted ``NPL'' column. For other datasets, we also observe that certain labels are too general and accommodate other labels during classification. We further observe that \textit{GPT-4o} excels at predicting correct subsets of true labels. 


\begin{table}[t!]
\centering

\caption{Average cosine similarities (Stella) between labels and all other labels.}
\centering
\resizebox{0.20\columnwidth}{!}{%
\begin{tabular}{c|c}
\textbf{Label Name} & \textbf{Cos. Sim.} \\ \hline
anger          & 0.66     \\ \hline
anticipation   &\textbf{0.72}     \\ \hline
disgust        & 0.67     \\ \hline
fear           & 0.69     \\ \hline
joy            & 0.70     \\ \hline
love           & 0.69     \\ \hline
optimism       & 0.69     \\ \hline
pessimism      & 0.69     \\ \hline
sadness        & 0.68     \\ \hline
surprise       & 0.67     \\ \hline
trust          & 0.67     \\ 
\end{tabular}%
}
\label{tab:sim-other-labels}
\end{table}

In general, we conclude that the introduced false positives can be attributed to the representations of texts and labels from sentence transformers, which seem to struggle with capturing the general semantics of a text. This causes the overlap in distributions, as discussed in Section~\ref{sec:distributions}. This issue is exemplified in a misclassified instance from the Reuters dataset:

\begin{tcolorbox}[
    colback=white,
    colframe=white,
    boxrule=0mm,
    sharp corners,
    left=0.5mm,
    right=0.5mm,
    top=0.5mm,
    bottom=0.5mm,
    fontupper=\footnotesize
]
\textbf{Text}: Shr loss five cts vs profit 10 cts // Net loss 381,391 vs profit 736,974 // Revs 6,161,391 vs 9,241,882 // NOTE: Canadian dollars. // Proved oil reserves at year-end 3.3 mln barrels, up 39 pct from a year earlier, and natural gas reserves 4.7 billion cubic feet, off nine pct.\\
\textbf{True}: Earnings and Forecasts. \textbf{Pred.}: Canadian Dollar, Crude Oil, Earnings and Forecasts, Natural Gas
\end{tcolorbox}

The reason the model predicts these labels is likely due to the presence of (partial) label names in the text. However, since these labels are not annotated, they are regarded as false positives, thereby mirroring the findings from~\cite{kosar-etal-2023-advancing}. Sentence transformers only manage to encode shallow semantics of the texts and fail to capture their general meaning.

Another false positive is observed in the following example, where a label name partially occurs in the text itself (``South-African''). However, the actual true label is only implied, which is challenging for distance-based models:

\begin{tcolorbox}[
    colback=white,
    colframe=white,
    boxrule=0mm,
    sharp corners,
    left=0.5mm,
    right=0.5mm,
    top=0.5mm,
    bottom=0.5mm,
    fontupper=\footnotesize
]
\textbf{Text}: Six black miners have been killed and two injured in a rock fall three km underground at a South African gold mine, the owners said on Sunday. Rand Mines Properties Ltd, one of South Africa's big six mining companies [...] \\
\textbf{True}: Gold. \textbf{Preds.}: South African Rand
\end{tcolorbox}

A similar mistake is observed in the SemEval dataset, where the model predicts ``anger'' for the following tweet due to the presence of the word ``furious'', which is actually part of a movie title:

\begin{tcolorbox}[
    colback=white,
    colframe=white,
    boxrule=0mm,
    sharp corners,
    left=0.5mm,
    right=0.5mm,
    top=0.5mm,
    bottom=0.5mm,
    fontupper=\footnotesize
]
\textbf{Text}: Fast and furious marathon soon! \\
\textbf{True}: anticipation, joy, optimism. \textbf{Pred.}: anger, anticipation, joy
\end{tcolorbox}

These mistakes highlight that the models are sensitive to context and (partial) matches of label names in texts, which may explain the higher performance of both fine-tuned models and LLMs on some datasets.

\section{Conclusion}
In this paper we investigated label-specific thresholds for distance-based MLTC, an under-explored yet computationally efficient classification method that is particularly well-suited for scenarios where there is insufficient data to train custom embeddings and classifiers. We first researched the possibility of label-specific thresholds by experimenting with several state-of-the-art sentence encoders on multiple publicly available datasets, and analyzed cosine similarity distributions. We found statistically significant differences between models (inter-model), datasets (intra-model), and labels (intra-model). These findings challenge the applicability of uniform thresholds for classification, and to a certain extent, for information retrieval tasks. We further applied these findings to classification experiments where we optimized thresholds based on annotated validation sets. The results showed that label-specific thresholds clearly outperform uniform thresholds and (normalized) 0.5, even when optimized on smaller data samples. Moreover, our method matches or even surpasses the performance of zero-shot LLMs on certain datasets.  

\section{Limitations and Future Work}
This study is subject to a few limitations. First and foremost, while our proposed method yields promising results, it still falls behind LLMs and fine-tuned models on some datasets, with the largest performance gap observed on the Reuters dataset. Second, the proposed label representation method of averaging keyword embeddings yields inconsistent results, which limits its applicability to other datasets. 

In addition, because of the black-box nature of the employed sentence encoders and LLMs, it cannot be determined whether the datasets have already been seen by the models during their pre-training phase. The presented results might therefore represent an overestimation of their performance. Nonetheless, our extensive evaluation of a multitude of models and sentence encoders trained from scratch has shown the effectiveness of our thresholding approach.

The performance of our thresholding approach is mainly limited by the separation between distributions in Figure~\ref{fig:dists-litcovid} and~\ref{fig:dists-biotech}. Better label representations or training regimes might encourage the models to more clearly separate similarity distributions between labels, therefore leading to better classification performance. Future work could explore more label representations or the use of contrastive learning to improve the representations of texts and labels. Additionally, we found that DBC is sensitive to (partial) matches of label names in texts, which can negatively affect the performance. Future work could focus on improving the encoding to capture deeper semantics, potentially overcoming this limitation.

\section{Acknowledgments}
This research was funded by Flanders Innovation \& Entrepreneurship (VLAIO), grant HBC.2021.0222 and by the Flemish government under FWO IRI project CLARIAH-VL.

\clearpage

\appendix
\renewcommand{\thesection}{Appendix~\Alph{section}}

\ifSubfilesClassLoaded{\appendix}{}

\section{Similarity Scale Variation Across Models (Exploratory Study)} \label{app:hyp1}

Table \ref{tab:mean-median-comp} contains the (normalized) mean and median cosine similarity scores for all models across all datasets.

\begin{table*}[h]
\caption{(Normalized) Mean and Median Cosine Similarity per Model and Dataset}
\centering
\resizebox{1\textwidth}{!}{%
\begin{tabular}{l|cccc|cccc|cccc|cccc|cccc}
\multicolumn{1}{l|}{} & 
\multicolumn{4}{c|}{\textbf{SemEval}} & 
\multicolumn{4}{c|}{\textbf{BioTech}} & 
\multicolumn{4}{c|}{\textbf{Reuters}} & 

\multicolumn{4}{c|}{\textbf{AAPD}} & 
\multicolumn{4}{c}{\textbf{LitCOVID}} \\ \hline

\multicolumn{1}{l|}{\textbf{model}} & 
\multicolumn{1}{c}{mn} & \multicolumn{1}{c}{\begin{tabular}[c]{@{}c@{}}mn\\(norm)\end{tabular}} & 
\multicolumn{1}{c}{mdn} & \multicolumn{1}{c|}{\begin{tabular}[c]{@{}c@{}}mdn\\(norm)\end{tabular}} & 

\multicolumn{1}{c}{mn} & \multicolumn{1}{c}{\begin{tabular}[c]{@{}c@{}}mn\\(norm)\end{tabular}} & 
\multicolumn{1}{c}{mdn} & \multicolumn{1}{c|}{\begin{tabular}[c]{@{}c@{}}mdn\\(norm)\end{tabular}} & 

\multicolumn{1}{c}{mn} & \multicolumn{1}{c}{\begin{tabular}[c]{@{}c@{}}mn\\(norm)\end{tabular}} & 
\multicolumn{1}{c}{mdn} & \multicolumn{1}{c|}{\begin{tabular}[c]{@{}c@{}}mdn\\(norm)\end{tabular}} &

\multicolumn{1}{l}{mn} & \multicolumn{1}{l}{\begin{tabular}[c]{@{}c@{}}mn\\(norm)\end{tabular}} & 
\multicolumn{1}{l}{mdn} & \multicolumn{1}{l|}{\begin{tabular}[c]{@{}c@{}}mdn\\(norm)\end{tabular}} & 

\multicolumn{1}{l}{mn} & \multicolumn{1}{l}{\begin{tabular}[c]{@{}c@{}}mn\\(norm)\end{tabular}} & 
\multicolumn{1}{l}{mdn} & \multicolumn{1}{l}{\begin{tabular}[c]{@{}c@{}}mdn\\(norm)\end{tabular}} \\ \hline

GIST-Large & 0.36 & 0.31 & 0.35 & 0.30 & 0.32 & 0.38 & 0.32 & 0.38 & 0.35 & 0.29 & 0.34 & 0.37 & 0.34 & 0.37 & 0.33 & 0.33 & 0.32 & 0.33 & 0.31 & 0.29 \\

Gte-Large & 0.48 & 0.44 & 0.48 & 0.44 & 0.45 & 0.42 & 0.45 & 0.43 & 0.36 & 0.43 & 0.35 & 0.46 & 0.43 & 0.48 & 0.43 & 0.41 & 0.44 & 0.41 & 0.43 & 0.44 \\

Stella & 0.35 & 0.25 & 0.34 & 0.24 & 0.28 & 0.24 & 0.28 & 0.23 & 0.26 & 0.26 & 0.25 & 0.25 & 0.33 & 0.35 & 0.32 & 0.35 & 0.33 & 0.34 & 0.31 & 0.27 \\

UAE-Large & 0.46 & 0.34 & 0.46 & 0.34 & 0.41 & 0.42 & 0.41 & 0.42 & 0.38 & 0.38 & 0.37 & 0.44 & 0.41 & 0.51 & 0.40 & 0.46 & 0.36 & 0.46 & 0.36 & 0.39 \\

Mxbai & 0.48 & 0.38 & 0.48 & 0.37 & 0.41 & 0.42 & 0.41 & 0.42 & 0.40 & 0.38 & 0.40 & 0.46 & 0.40 & 0.49 & 0.39 & 0.46 & 0.37 & 0.46 & 0.37 & 0.39 \\

BGE & 0.50 & 0.44 & 0.50 & 0.44 & 0.46 & 0.40 & 0.46 & 0.40 & 0.35 & 0.43 & 0.34 & 0.45 & 0.40 & 0.53 & 0.40 & 0.52 & 0.41 & 0.52 & 0.41 & 0.43 \\

SF & 0.44 & 0.42 & 0.45 & 0.42 & 0.31 & 0.41 & 0.31 & 0.40 & 0.34 & 0.30 & 0.33 & 0.34 & 0.42 & 0.41 & 0.41 & 0.35 & 0.36 & 0.35 & 0.36 & 0.30 \\

all-MPNet & 0.13 & 0.30 & 0.12 & 0.29 & 0.14 & 0.36 & 0.13 & 0.35 & 0.29 & 0.07 & 0.27 & 0.22 & 0.31 & 0.10 & 0.30 & 0.12 & 0.33 & 0.11 & 0.32 & 0.09 \\

all-MPNet-Wiki & 0.34 & 0.33 & 0.33 & 0.32 & 0.32 & 0.43 & 0.32 & 0.42 & 0.34 & 0.23 & 0.32 & 0.38 & 0.47 & 0.39 & 0.46 & 0.41 & 0.42 & 0.41 & 0.42 & 0.25 \\

GloVe & 0.23 & 0.39 & 0.22 & 0.38 & 0.45 & 0.60 & 0.46 & 0.60 & 0.45 & 0.19 & 0.44 & 0.47 & 0.60 & 0.47 & 0.61 & 0.48 & 0.65 & 0.48 & 0.65 & 0.21 \\

\end{tabular}%
}
\label{tab:mean-median-comp}
\end{table*}

Figure \ref{fig:norm_domain_variability} shows the variability of cosine similarity scales across models and datasets after normalization.

\begin{figure}[ht!]
\centering
\includegraphics[width=0.8\linewidth]{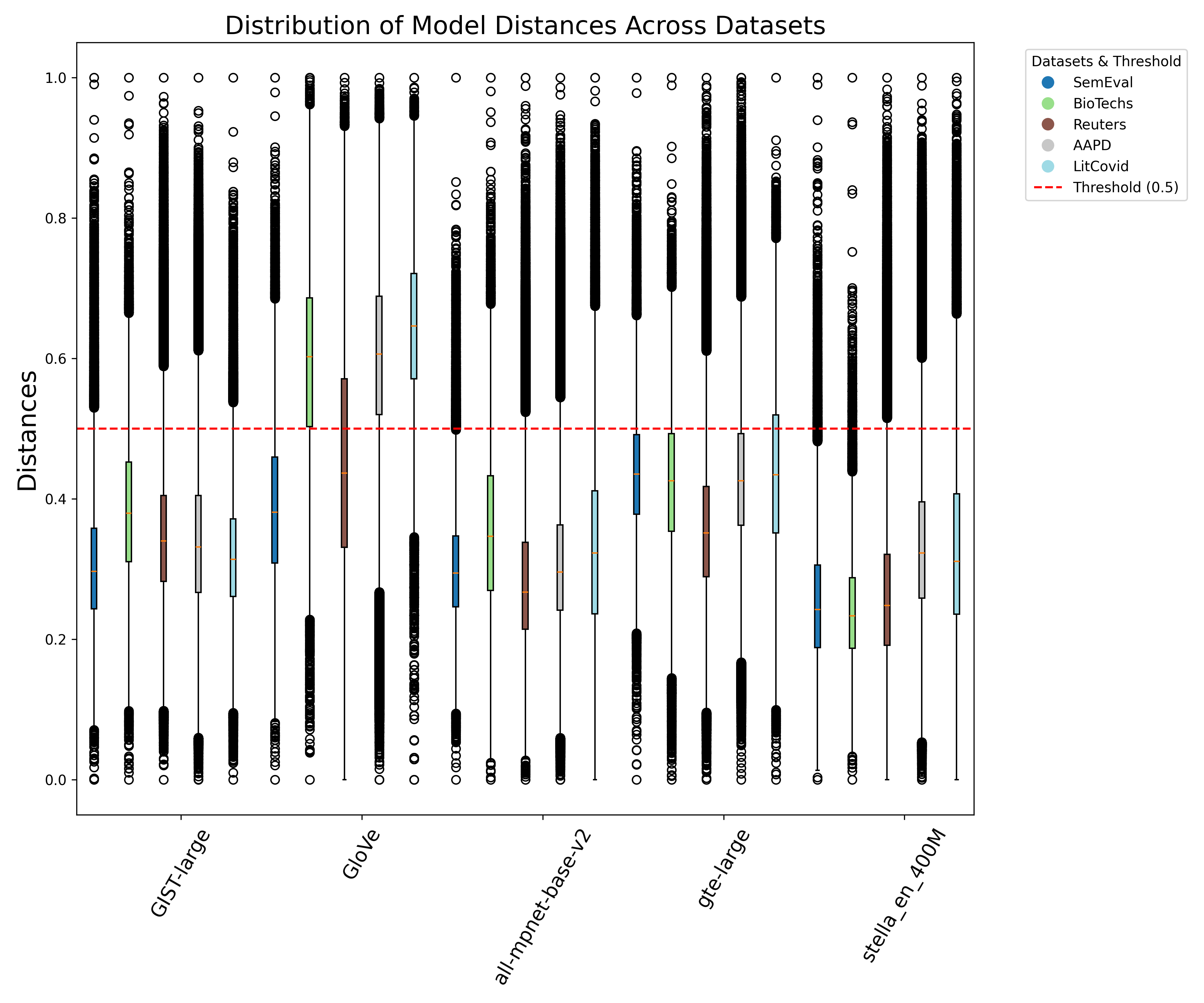}
\caption{Bar charts showing the variation in normalized cosine similarity distributions between individual models and domains.}
\label{fig:norm_domain_variability}
\end{figure}

\clearpage
\section{Computation Times} \label{app:times}

The total computation time, calculated as the sum of fine-tuning or threshold calibration and inference process across all datasets, is summarised in Table \ref{tab:times}. It should be noted that, though iterating over thresholds for uniform thresholds is evidently much faster, more computational resources are spent on calculating the macro-averaged $F_1$ score for each threshold as opposed to the $F_1$ score of the positive class for label-specific thresholds. The batch size (8) remained consistent across all experiments.

\begin{table}[!h]
\small
\centering
\caption{Sum of Training + Inference Times (in hours) across Datasets, with Consistent Batch Size (8).}
\centering
\resizebox{0.5\columnwidth}{!}{%
\begin{tabular}{r|c|c|c}
\textbf{Model} & \textbf{Lbl. rep.} & \textbf{Thresh.} & \textbf{\begin{tabular}[c] {@{}r@{}}Train + Inf.\\ (hrs)\end{tabular}} \\ \hline
RoBERTa (sample) & / & /& 5.44 \\
Stella & lbl. name & lbl. & 0.24 \\
Stella & kw & lbl. & 0.25 \\
Stella & kw & unif. & 0.26
\end{tabular}%
}
\label{tab:times}
\end{table}

\section{Evaluation of Label Representations} \label{app:label-adj} 

Table \ref{tab:lbl-name-comp-datasets} compares the performance of different label representations methods (out-of-the-boxlabel names and adjusted label names). The results are obtained using the best performing model for each dataset. Table \ref{tab:lbl-name-comp-litcovid} shows a per-class comparison between the two representation methods for the LitCovid dataset. Table \ref{tab:results-label-rep} compares the results with label names and generated keywords as label representations.

\begin{table}[ht!]
\caption{Comparison between the Original Label Names and Adjusted Label Names (Best Model for each Dataset).}
\centering
\resizebox{0.4\columnwidth}{!}{%
\begin{tabular}{l|cc|cc}
 & \multicolumn{2}{l|}{\textbf{Lbl Names}} & \multicolumn{2}{l}{\textbf{Adj. Lbl Names}} \\
\textbf{Dataset} & \multicolumn{1}{c}{ma$F_1$} & \multicolumn{1}{c|}{mi$F_1$} & \multicolumn{1}{c}{ma$F_1$} & \multicolumn{1}{c}{mi$F_1$} \\ \hline
BioTech & 24.36 & 36.86 & 27.15 & 36.97 \\
Reuters & 46.53 & 60.79 & 47.63 & 62.69 \\
LitCovid & 54.82 & 58.07 & 61.68 & 58.30
\end{tabular}%
}
\label{tab:lbl-name-comp-datasets}
\end{table}

\begin{table}[!h]
\centering
\caption{Comparison between the Original Label Names and Adjusted Label Names on the LitCovid dataset (Stella)}
\label{tab:lbl-name-comp-litcovid}
\resizebox{0.7\columnwidth}{!}{%
\begin{tabular}{rccc|rccc}
\textbf{Label Name}  & \textbf{pr}    & \textbf{rec}   & \textbf{F1}    & \textbf{Adj. Label Name} & \textbf{pr}    & \textbf{rec}   & \textbf{F1}     \\ 
\hline
Case Report          & 45.51          & 59.96          & 51.75          & Case Report              & 45.51          & 59.96          & 51.75           \\
Diagnosis            & 56.37          & \textbf{72.70} & 63.50          & Diagnostic Methods       & \textbf{64.29} & 65.91          & \textbf{65.09}  \\
Epidemic Forecasting & 59.26          & 75.00          & 66.21          & Epidemic Forecasting     & 59.26          & 75.00          & 66.21           \\
Mechanism            & 27.43          & 54.52          & 36.49          & Biological Mechanisms    & \textbf{58.99} & \textbf{70.92} & \textbf{64.41}  \\
Prevention           & 61.26          & \textbf{80.15} & 69.44          & Prevention Strategies    & \textbf{67.90} & 79.67          & \textbf{73.31}  \\
Transmission         & \textbf{33.88} & \textbf{48.05} & \textbf{39.74} & Transmission Dynamics    & 27.32          & 43.75          & 33.63           \\
Treatment            & \textbf{49.86} & 65.52          & \textbf{56.63} & Medical Treatments       & 38.28          & \textbf{90.03} & 53.72           \\ 
\hline
micro avg            & 49.85          & 69.54          & 58.07          & micro avg                & \textbf{51.69} & \textbf{76.45} & \textbf{61.68}  \\
macro avg            & 47.65          & 65.13          & 54.82          & macro avg                & \textbf{51.65} & \textbf{69.32} & \textbf{58.30} 
\end{tabular}%
}
\end{table}

\begin{table*}[!h]
\caption{Comparison between Label Names and Keyword Embeddings as Label Representations for Optimized Label-Specific Thresholds.}
\centering
\resizebox{1\columnwidth}{!}{%
\begin{tabular}{l|r|rrr|rrr|rrr|rrr|rrr}
\multicolumn{1}{c}{\textbf{}} & 
\textbf{} & \multicolumn{3}{c|}{\textbf{SemEval}} & 
\multicolumn{3}{c|}{\textbf{BioTech}} & 
\multicolumn{3}{c|}{\textbf{Reuters}} & 

\multicolumn{3}{c|}{\textbf{AAPD}} & 
\multicolumn{3}{c}{\textbf{LitCovid}} \\ \hline
\multicolumn{1}{l|}{\textbf{model}} & \textbf{rep.} & 
\multicolumn{1}{l}{\textbf{ma$F_1$}} & \multicolumn{1}{l}{\textbf{mi$F_1$}} & \multicolumn{1}{l|}{\textbf{$P@1$}} & 
\multicolumn{1}{l}{\textbf{ma$F_1$}} & \multicolumn{1}{l}{\textbf{mi$F_1$}} & \multicolumn{1}{l|}{\textbf{$P@1$}} & 
\multicolumn{1}{l}{\textbf{ma$F_1$}} & \multicolumn{1}{l}{\textbf{mi$F_1$}} & \multicolumn{1}{l|}{\textbf{$P@1$}} & 

\multicolumn{1}{l}{\textbf{ma$F_1$}} & \multicolumn{1}{l}{\textbf{mi$F_1$}} & \multicolumn{1}{l|}{\textbf{$P@1$}} & 
\multicolumn{1}{l}{\textbf{ma$F_1$}} & \multicolumn{1}{l}{\textbf{mi$F_1$}} & \multicolumn{1}{l}{\textbf{$P@1$}} \\ \hline

\multirow{2}{*}{GIST-Large} & lbl. & 
45.15 & 53.37 & \textbf{68.18} & 
22.56 & 32.48 & 20.73 & 
\textbf{31.59} & \textbf{38.02} & 50.55 & 

41.06 & \textbf{49.97} & 64.3 & 
\textbf{51.34} & 56.31 & 44.96 \\

& kw & 
\textbf{47.59} & \textbf{55.00} & 67.51 & 
\textbf{23.29} & \textbf{32.99} & \textbf{28.08} & 
29.31 & 36.63 & \textbf{52.10} & 

\textbf{41.78} & 46.84 & \textbf{64.7} & 
48.31 & \textbf{56.76} & \textbf{61.07} \\ \hline

\multirow{2}{*}{GTE-Large} & lbl. & 
41.20 & 49.85 & \textbf{63.42} & 
\textbf{26.18} & \textbf{36.63} & 26.77 & 
\textbf{34.47} & \textbf{41.31} & 48.76 & 

35.95 & 40.24 & 61.0 & 
\textbf{48.79} & \textbf{55.71} & 51.16 \\

& kw & 
\textbf{44.01} & \textbf{50.87} & 61.95 & 
24.13 & 35.65 & \textbf{26.77} & 
30.39 & 41.10 & \textbf{48.86} & 

\textbf{37.41} & \textbf{40.13} & \textbf{62.0} & 
47.76 & 55.14 & \textbf{60.65} \\ \hline

\multirow{2}{*}{Stella} & lbl. & 
41.20 & \textbf{48.96} & \textbf{65.91} & 
\textbf{26.10} & 34.45 & 23.10 & 
\textbf{37.14} & \textbf{55.54} & 69.04 & 

41.65 & 46.60 & 56.8 & 
\textbf{58.30} & 61.68 & 18.67 \\
& kw & 
\textbf{42.54} & 48.66 & 61.83 & 
26.02 & \textbf{35.27} & \textbf{28.87} & 
33.53 & 47.20 & \textbf{69.24} & 

\textbf{42.41} & \textbf{48.16} & \textbf{64.1} & 
57.71 & \textbf{63.88} & \textbf{55.47} \\ \hline

\multirow{2}{*}{UAE-Large} & lbl. & 
\textbf{44.05} & \textbf{53.07} & \textbf{64.34} & 
\textbf{22.55} & \textbf{32.49} & 22.57 & 
\textbf{34.01} & 35.37 & 46.44 & 

36.40 & \textbf{44.24} & 62.5 & 
\textbf{45.77} & \textbf{53.89} & 39.77 \\

 & kw &
43.96 & 51.65 & 63.82 & 
22.42 & 31.78 & \textbf{25.20} & 
28.30 & \textbf{37.08} & \textbf{48.03} & 

\textbf{38.60} & 43.86 & \textbf{64.0} & 
42.11 & 53.05 & \textbf{56.61} \\ \hline

\multirow{2}{*}{Mxbai} & lbl. & 
\textbf{44.42} & \textbf{54.52} & 64.50 & 
21.74 & 31.03 & 23.88 & 
\textbf{32.88} & 33.84 & 47.14 & 

38.84 & \textbf{47.18} & 62.8 & 
\textbf{48.48} & \textbf{54.97} & 42.15 \\

& kw &
43.93 & 50.82 & \textbf{64.68} & 
\textbf{25.23} & \textbf{33.75} & \textbf{27.82} & 
28.10 & \textbf{34.32} & \textbf{50.88} & 

\textbf{39.79} & 44.92 & \textbf{63.6} & 
45.44 & 54.94 & \textbf{59.22} \\ \hline

\multirow{2}{*}{BGE} & lbl. & 
42.10 & 48.62 & 57.32 & 
19.72 & \textbf{30.75} & 17.85 & 
\textbf{32.97} & \textbf{43.43} & 51.04 & 

34.77 & 40.61 & 49.6 & 
\textbf{48.08} & \textbf{56.53} & 53.55 \\

& kw & 
\textbf{42.68} & \textbf{48.97} & \textbf{61.64} &
\textbf{21.63} & 30.41 & \textbf{21.52} & 
26.07 & 36.85 & \textbf{55.44} & 

\textbf{37.05} & \textbf{43.97} & \textbf{61.4} & 
45.75 & 54.27 & \textbf{62.40} \\ \hline

\multirow{2}{*}{SF} & lbl. & 
\textbf{42.31} & \textbf{53.20} & 61.46 & 
\textbf{22.16} & 31.12 & 16.54 & 
\textbf{32.74} & \textbf{43.03} & 45.05 & 

\textbf{33.73} & 40.75 & \textbf{66.2} & 
\textbf{49.19} & 56.56 & 35.60 \\

& kw & 
41.03 & 50.41 & \textbf{64.41} & 
21.02 & \textbf{32.51} & \textbf{25.46} & 
28.93 & 42.65 & \textbf{51.21} & 

33.26 & \textbf{43.08} & 63.1 & 
47.22 & \textbf{56.99} & \textbf{61.82} \\ \hline

\multirow{2}{*}{all-MPNet} & lbl. & 
38.73 & 49.77 & \textbf{59.71} & 
20.71 & 25.37 & 17.85 & 
\textbf{30.49} & \textbf{41.76} & 49.72 & 

37.06 & 41.95 & 62.1 & 
\textbf{53.41} & \textbf{59.98} & 19.79 \\

 & kw & 
\textbf{40.59} & \textbf{51.91} & 57.04 & 
\textbf{22.26} & \textbf{26.07} & \textbf{22.05} & 
27.52 & 39.29 & \textbf{59.21} & 

\textbf{39.70} & \textbf{46.95} & \textbf{66.9} & 
51.32 & 56.68 & \textbf{46.91} \\ \hline

\multirow{2}{*}{\begin{tabular}[c]{@{}r@{}}all-MPNet-Wiki\end{tabular}} & lbl. & 
37.90 & 46.13 & 51.40 & 
21.04 & 26.86 & 19.69 & 
\textbf{28.57} & \textbf{48.70} & \textbf{61.40} & 

37.38 & 45.11 & 61.2 & 
49.37 & 57.36 & 17.17 \\

 & kw & 
\textbf{38.65} & \textbf{46.87} & \textbf{56.55} & 
\textbf{22.62} & \textbf{27.95} & \textbf{19.69} & 
25.48 & 39.80 & 60.77 & 

\textbf{39.43} & \textbf{45.13} & \textbf{65.0} & 
\textbf{51.14} & \textbf{58.00} & \textbf{37.81} \\ \hline

\multirow{2}{*}{GloVe} & lbl. & 
36.22 & 42.53 & 28.69 & 
20.45 & 28.46 & 7.61 & 
\textbf{18.70} & \textbf{23.88} & 28.48  &

19.00 & 25.14 & 12.0 &
38.43 & 44.26 & \textbf{43.31} \\

 & kw & 
\textbf{37.17} & \textbf{44.18} & \textbf{31.91} & 
\textbf{20.94} & \textbf{30.97} & \textbf{10.24} & 
15.90 & 23.05 & \textbf{29.90} & 

\textbf{22.07} & \textbf{30.00} & \textbf{32.3} & 
\textbf{44.78} & \textbf{56.67} & 28.71 

\end{tabular}%
 }

\label{tab:results-label-rep}
\end{table*}

\clearpage

\section{Complete Classification Results}\label{app:comp-results}

Table \ref{tab:all-classification-results} contains the classification results for all thresholding methods for all models.

\begin{table*}[!ht]
\centering
\caption{Classification Results on All Datasets. An Asterisk (*) Indicates Best Performance with Label Names.}
\label{tab:all-classification-results}
\resizebox{1\textwidth}{!}{%
\begin{tabular}{l|l|lll|lll|lll|lll|lll}
\multicolumn{1}{l}{}& \multicolumn{1}{c|}{} & 
\multicolumn{3}{c|}{\textbf{SemEval}}  & 
\multicolumn{3}{c|}{\textbf{BioTech}} & 
\multicolumn{3}{c|}{\textbf{Reuters}}  & 

\multicolumn{3}{c|}{\textbf{AAPD}} & 
\multicolumn{3}{c}{\textbf{LitCovid}} \\ 

\hline
\multicolumn{1}{l|}{\textbf{model}}& \multicolumn{1}{c|}{\textbf{thr}} & 
\multicolumn{1}{c}{\textbf{ma$F_1$}}& \multicolumn{1}{c}{\textbf{mi$F_1$}}  & \multicolumn{1}{c|}{\textbf{$P@1$}} & 
\multicolumn{1}{c}{\textbf{ma$F_1$}} & \multicolumn{1}{c}{\textbf{mi$F_1$}}& \multicolumn{1}{c|}{\textbf{$P@1$}} & 
\multicolumn{1}{c}{\textbf{ma$F_1$}}& \multicolumn{1}{c}{\textbf{mi$F_1$}} & \multicolumn{1}{c|}{\textbf{$P@1$}} & 

\multicolumn{1}{c}{\textbf{ma$F_1$}}& \multicolumn{1}{c}{\textbf{mi$F_1$}} & \multicolumn{1}{c|}{\textbf{$P@1$}} & 
\multicolumn{1}{c}{\textbf{ma$F_1$}}& \multicolumn{1}{c}{\textbf{mi$F_1$}}& \multicolumn{1}{c}{\textbf{$P@1$}}\\ 
\hline

\multirow{4}{*}{GIST-Large}
& 0.5 & 42.2  & 48.58& 67.51 & 18.6& 19.3  & 28.1  & 32.8  & 37.92  & 50.55  & 28.46 & 30.02  & 64.7  & 35.35 & 42.95 & 61.07  \\
& n0.5& 42.94 & 49.53& 67.51 & 15.43  & 19.0  & 28.1  & 12.3  & 15.81  & 50.55  & 30.89 & 32.64  & 64.7  & 40.76 & 48.47 & 61.07  \\
& unif& 44.99 & 51.39& 67.51 & 18.54  & 21.5  & 28.1  & \textbf{32.8}  & 37.92  & 50.55  & 38.21 & 43.11  & 64.7  & 40.32 & 47.41 & 61.07  \\
& lbl & {\cellcolor[rgb]{0.937,0.937,0.937}}\textbf{47.59} & {\cellcolor[rgb]{0.937,0.937,0.937}}\textbf{55.0} & {\cellcolor[rgb]{0.937,0.937,0.937}}\textbf{67.51} & \textbf{23.29}  & \textbf{33.0}  & \textbf{28.1}  & 31.6* & \textbf{38.02*} & \textbf{50.55*} & \textbf{41.06*}& {\cellcolor[rgb]{0.937,0.937,0.937}}\textbf{49.97*} & {\cellcolor[rgb]{0.937,0.937,0.937}}\textbf{64.3*} & \textbf{51.34*}& \textbf{56.31*}& \textbf{44.96*} \\ 

\hline
\multirow{4}{*}{GTE-Large}
& 0.5 & 37.25 & 39.42& 61.95 & 13.95  & 16.7  & 26.8  & 8.3& 11.44  & 48.76  & 17.51*& 18.91* & 61.0* & 38.82 & 44.07 & 60.65  \\
& n0.5& 41.23 & 45.84& 61.95 & 13.32  & 15.9  & 26.8  & 8.86  & 12.2& 48.76  & 23.98 & 27.02  & 62.0  & 39.76 & 47.55 & 60.65  \\
& unif& 41.35 & 46.79& 61.95 & 19.56  & 21.0  & 26.8  & 33.2* & 38.13* & 48.76* & 33.88 & 39.49  & 62.0  & 43.64*& 48.53*& 51.16* \\
& lbl & \textbf{44.01} & \textbf{50.87}& \textbf{61.95} & {\cellcolor[rgb]{0.937,0.937,0.937}}\textbf{26.18*} & {\cellcolor[rgb]{0.937,0.937,0.937}}\textbf{36.6*} & {\cellcolor[rgb]{0.937,0.937,0.937}}\textbf{26.8*} & \textbf{34.5*} & \textbf{41.31*} & \textbf{48.76*} & \textbf{37.41} & \textbf{40.13}  & \textbf{62.0}  & \textbf{48.79*}& \textbf{55.71*}& \textbf{51.16*} \\ 

\hline
\multirow{4}{*}{Stella}
& 0.5 & 28.91 & 32.37& 61.83 & 6.29& 2.34  & 28.9  & 29.2  & 29.41  & 69.04  & 35.19 & 43.68  & 64.1  & 29.2  & 28.36 & 55.47  \\
& n0.5& 25.62 & 28.45& 61.83 & 18.03  & 19.9  & 28.9  & 20.9  & 29.83  & 69.04  & 31.81*& 35.07* & 56.8* & 43.68 & 46.43 & 55.47  \\
& unif& 41.54 & 45.9 & 61.83 & 18.25  & 19.4  & 28.9  & 36.5* & 54.26* & 69.04* & 35.19 & 43.68  & 64.1  & 44.5  & 47.4  & 55.47  \\
& lbl & \textbf{42.54} & \textbf{48.66}& \textbf{61.83} & \textbf{26.02}  & \textbf{35.3}  & \textbf{28.9}  & {\cellcolor[rgb]{0.937,0.937,0.937}}\textbf{37.1*} & {\cellcolor[rgb]{0.937,0.937,0.937}}\textbf{55.54*} & {\cellcolor[rgb]{0.937,0.937,0.937}}\textbf{69.04*} & {\cellcolor[rgb]{0.937,0.937,0.937}}\textbf{42.41} & \textbf{48.16}  & \textbf{64.1}  & {\cellcolor[rgb]{0.937,0.937,0.937}}\textbf{57.71} & {\cellcolor[rgb]{0.937,0.937,0.937}}\textbf{63.88} & \textbf{55.47}  \\ 

\hline

\multirow{4}{*}{UAE-Large}
& 0.5& 42.08*& 48.1*& 64.34*& 16.85  & 20.7  & 25.2  & 22.7  & 27.39  & 46.44 & 12.85*& 14.09* & 62.5* & 31.44 & 33.96 & 56.61  \\
& n0.5& 40.98 & 45.65& 63.82 & 15.15  & 18.4  & 25.2  & 12.1  & 16.0& 46.44  & 21.78 & 22.68  & 64.0  & 35.58 & 41.7  & 56.61  \\
& unif& 42.1  & 48.69& 63.82 & 17.95  & 20.8  & 25.2  & 31.4* & 32.4*  & 46.44* & 33.28 & 38.15  & 64.0  & 35.36 & 40.97 & 56.61  \\
& lbl& \textbf{44.05*}& \textbf{53.07*}  & \textbf{64.34*}& \textbf{22.55}  & \textbf{32.49} & \textbf{25.2}  & \textbf{34.0*} & \textbf{35.37*} & \textbf{46.44*} & \textbf{38.6}  & \textbf{43.86}  & \textbf{64.0}  & \textbf{45.77*}& \textbf{53.89*}& \textbf{39.77*} \\ 

\hline

\multirow{4}{*}{Mxbai}
& 0.5& 42.89*& 47.76*  & 64.5* & 17.91* & 19.31*& 23.88*& 22.14*& 24.85* & 47.14* & 15.35*& 16.67* & 62.8* & 33.29*& 35.67*& 42.15* \\
& n0.5& 39.86 & 46.43& 64.5  & 15.79  & 19.34 & 27.82 & 9.98* & 11.62* & 47.14* & 23.5  & 24.44  & 63.6  & 37.33 & 43.37 & 59.22  \\
& unif& 42.89 & 49.16& 64.68 & 20.2& 22.7  & 27.8  & 30.6* & \textbf{35.08*} & 47.14* & 34.99 & 39.72  & 63.6  & 38.81*& 40.49*& 42.15* \\
& lbl& \textbf{44.42*}& \textbf{54.52*}  & \textbf{64.50*}& \textbf{25.23}  & \textbf{33.8}  & \textbf{27.8}  & \textbf{32.9*} & 33.84* & \textbf{47.14*} & \textbf{39.79} & \textbf{44.92}  & \textbf{63.6}  & \textbf{48.48*}& \textbf{54.97*}& \textbf{42.15*} \\ 

\hline

\multirow{4}{*}{BGE}
& 0.5& 39.51*& 42.56*  & 57.32*& 15.57* & 17.18*& 17.85*& 11.4* & 13.34  & 51.04  & 11.03*& 11.69* & 49.0* & 34.94*& 38.4* & 53.55* \\
& n0.5  & 40.35*& 43.83*  & 57.32*& 15.13* & 17.3* & 17.85*& 8.36  & 9.95& 51.04  & 23.39 & 24.09  & 49.6  & 35.4* & 39.3* & 53.55* \\
& unif& 39.31 & 44.93& 61.64 & 16.77  & 19.5  & 21.5  & 32.1* & 35.82* & 51.04* & 31.62 & 36.64  & 61.4  & 40.53*& 45.24*& 53.55* \\
& lbl & \textbf{42.68} & \textbf{48.97}& \textbf{61.64} & \textbf{21.63}  & \textbf{30.4}  & \textbf{21.5}  & \textbf{33.0*} & \textbf{43.43*} & \textbf{51.04*} & \textbf{37.05} & \textbf{43.97}  & \textbf{61.4}  & \textbf{48.08*}& \textbf{56.53*}& \textbf{53.55*} \\ 

\hline

\multirow{4}{*}{SF}
& 0.5& 38.32*& 44.47*  & 61.46*& 15.06  & 12.7  & 25.5  & 32.1  & 31.11  & 45.05  & 27.54*& 32.99* & 66.2* & 36.87 & 44.21 & 61.82  \\
& n0.5  & 38.85*& 44.87*  & 61.46*& 15.33  & 19.8  & 25.5  & 15.0  & 18.81  & 45.05 & 18.3  & 20.52  & 66.2  & 39.08 & 46.47 & 61.82  \\
& unif& 41.29 & 47.77& 70.17 & 15.74  & 18.4  & 22.8  & 31.8* & 32.42* & 45.05* & 31.29 & 35.02  & 63.1  & 42.03 & 49.18 & 61.82  \\
& lbl& \textbf{42.31*}& \textbf{53.20*}  & \textbf{64.41*}& \textbf{20.74}  & \textbf{34.6}  & \textbf{22.8}  & \textbf{32.7*} & \textbf{43.03*} & \textbf{45.05*} & \textbf{33.26*}& \textbf{43.08}  & \textbf{63.1}  & \textbf{47.22} & \textbf{56.99} & {\cellcolor[rgb]{0.937,0.937,0.937}}\textbf{61.82}  \\

\hline

\multirow{4}{*}{all-MPNet}
& 0.5 & 1.38  & 1.64 & 57.04 & 5.96& 2.08  & 22.05 & 9.62* & 6.34*  & 49.72* & 10.46 & 10.33  & 66.9  & 8.64  & 4.64  & 46.91  \\
& n0.5& 29.16 & 33.88& 57.04 & 14.58  & 18.46 & 22.05 & 20.8* & 22.77* & 49.72* & 32.73 & 38.62  & 66.9  & 36.31 & 42.64 & 46.91  \\
& unif& 39.14 & 44.28& 57.04 & 12.59  & 16.0  & 22.05 & 27.2* & 34.12* & 49.52* & 35.88 & 43.13  & 66.9  & 41.0* & 43.94*& 20.12* \\
& lbl & \textbf{40.59} & \textbf{51.91}& \textbf{57.04} & \textbf{22.26}  & \textbf{26.07} & \textbf{22.05} & \textbf{30.6}  & \textbf{39.71}  & \textbf{49.52}  & \textbf{39.68} & \textbf{46.96}  & \textbf{66.9}  & \textbf{53.56} & \textbf{60.15} & \textbf{20.12}  \\ 

\hline

\multirow{4}{*}{all-MPNet-Wiki}
& 0.5 & 29.76 & 33.91& 56.55 & 16.67  & 19.07 & 19.69 & 22.14*& 31.68* & 61.4*  & 29.2  & 32.72  & 61.2  & 35.84 & 37.36 & 37.81  \\
& n0.5& 36.12 & 40.78& 56.55 & 15.22  & 17.11 & 19.69 & 9.85* & 12.7*  & 61.4*  & 20.29 & 21.34  & 61.2  & 36.66*& 41.65*& 17.17* \\
& unif& 37.08 & 41.04& 56.55 & 16.13  & 19.0  & 19.7  & 26.2* & 43.43* & 60.77* & 37.66 & 42.32  & 66.1  & 43.15*& 48.63*& 23.99* \\
& lbl & \textbf{38.65} & \textbf{46.87}& \textbf{56.55} & \textbf{22.62}  & \textbf{27.95} & \textbf{19.7}  & \textbf{28.3*} & \textbf{49.23*} & \textbf{60.77*} &  \textbf{39.63} & \textbf{47.46}  & \textbf{66.1}  & \textbf{53.05} & \textbf{60.46} & \textbf{43.69}  \\ 

\hline

\multirow{4}{*}{GloVe}
& 0.5 & 18.64 & 20.06& 31.91 & 11.44* & 14.52*& 7.61* & 10.3  & 9.58& 28.48  & 12.35*& 13.98* & 12 & 35.61*& 43.86*& 43.31* \\
& n0.5& 32.42 & 34.71& 31.91 & 12.73* & 14.62*& 7.61* & 5.97  & 5.26& 28.48  & 9.13  & 9.86& 32.3  & 30.54*& 34.22*& 43.31* \\
& unif& 34.25 & 35.94& 31.91 & 14.2& 16.0  & 10.2  & 9.27* & 7.38*  & 28.48* & 16.62 & 18.01  & 32.3  & 35.63*& 43.39*& 43.31* \\
& lbl & \textbf{37.17} & \textbf{44.18}& \textbf{31.91} & \textbf{20.94}  & \textbf{30.97} & \textbf{10.24} & \textbf{18.7*} & \textbf{23.88*} & \textbf{28.48*} & \textbf{22.07} & \textbf{30.0}& \textbf{32.3}  & \textbf{44.78} & \textbf{56.67} & \textbf{28.71}  \\ 

\end{tabular}%
}
\end{table*}

\clearpage

\section{Classification Results with Euclidean Distance}

Table \ref{tab:all-results-euclid} contains the results obtained using the Euclidean distance as a distance metric. 

\begin{table*}[!h]
\caption{Results with Euclidean Distance}
\label{tab:all-results-euclid}
\resizebox{1\textwidth}{!}{%
\begin{tabular}{l|l|rrr|rrr|rrr|rrr|rrr}
\multicolumn{1}{r|}{\textbf{}} &  & 

\multicolumn{3}{c|}{\textbf{SemEval}} & 
\multicolumn{3}{c|}{\textbf{BioTech}} & 
\multicolumn{3}{c|}{\textbf{Reuters}} & 

\multicolumn{3}{c|}{\textbf{AAPD}} & 
\multicolumn{3}{c}{\textbf{LitCovid}} \\ 

\hline

\multicolumn{1}{l|}{\textbf{model}} & \textbf{thr} & 
\multicolumn{1}{l}{\textbf{ma$F_1$}} & \multicolumn{1}{l}{\textbf{mi$F_1$}} & \multicolumn{1}{l|}{\textbf{$P@1$}} & 
\multicolumn{1}{l}{\textbf{ma$F_1$}} & \multicolumn{1}{l}{\textbf{mi$F_1$}} & \multicolumn{1}{l|}{\textbf{$P@1$}} & 
\multicolumn{1}{l}{\textbf{ma$F_1$}} & \multicolumn{1}{l}{\textbf{mi$F_1$}} & \multicolumn{1}{l|}{\textbf{$P@1$}} & 

\multicolumn{1}{l}{\textbf{ma$F_1$}} & \multicolumn{1}{l}{\textbf{mi$F_1$}} & \multicolumn{1}{l|}{\textbf{$P@1$}} & 
\multicolumn{1}{l}{\textbf{ma$F_1$}} & \multicolumn{1}{l}{\textbf{mi$F_1$}} & \multicolumn{1}{l}{\textbf{$P@1$}} \\ 

\hline

\multirow{2}{*}{GIST-Large}
& unif & 43.32 & 49.96 & 68.18 & 14.65 & 17.10 & 20.73 & \textbf{31.99} & 33.04 & 50.55 & 33.76 & 40.11 & 64.3 & 38.38 & 41.56 & 1.59 \\
& lbl & \textbf{45.29} & \textbf{54.16} & \textbf{68.18} & \textbf{17.00} & \textbf{22.41} & \textbf{20.73} & 27.96 & \textbf{37.91} & \textbf{50.55} & \textbf{40.01} & \textbf{46.73} & \textbf{64.3} & \textbf{41.31} & \textbf{51.26} & \textbf{1.59} \\ 

\hline

\multirow{2}{*}{GTE-Large}
& unif & 39.08 & 45.51 & 62.44 & 16.19 & 18.96 & 27.82 & \textbf{33.24} & 37.06 & 47.60 & 29.61 & 33.91 & 60.3 & 44.29 & 48.45 & 1.62 \\
& lbl & \textbf{39.82} & \textbf{45.27} & \textbf{62.44} & \textbf{19.41} & \textbf{26.37} & \textbf{27.82} & 33.00 & \textbf{44.38} & \textbf{47.60} & \textbf{30.89} & \textbf{34.52} & \textbf{60.3} & \textbf{48.76} & \textbf{55.68} & \textbf{1.62} \\ 

\hline

\multirow{2}{*}{Stella}
& unif & 40.60 & 42.46 & 52.68 & 11.78 & 12.93 & 20.21 & 31.05 & 52.38 & 72.28 & 31.63 & 35.15 & 47.9 & 35.58 & 36.95 & 6.84 \\
& lbl & \textbf{41.66} & \textbf{49.52} & \textbf{52.68} & \textbf{13.26} & \textbf{17.09} & \textbf{20.21} & \textbf{37.06} & \textbf{56.72} & \textbf{72.28} & \textbf{42.92} & \textbf{48.97} & \textbf{47.9} & \textbf{54.70} & \textbf{58.84} & \textbf{6.84} \\ 

\hline

\multirow{2}{*}{UAE-Large}
& unif & 41.49 & 46.50 & 62.29 & 12.28 & 14.34 & 26.51 & \textbf{31.28} & 31.04 & 36.95 & 29.81 & 30.60 & 49.1 & 35.87 & 39.79 & 3.41 \\
& lbl & \textbf{42.01} & \textbf{49.67} & \textbf{62.29} & \textbf{14.05} & \textbf{19.11} & \textbf{26.51} & 28.98 & \textbf{32.98} & \textbf{36.95} & \textbf{36.46} & \textbf{41.86} & \textbf{49.1} & \textbf{41.39} & \textbf{48.57} & \textbf{3.41} \\

\hline

\multirow{2}{*}{Mxbai}

& unif & 41.58 & 46.11 & 61.34 & 14.09 & 16.06 & 25.20 & \textbf{28.64} & 29.98 & 39.03 & 30.82 & 31.25 & 52.6 & 36.59 & 41.16 & 3.05 \\
& lbl & \textbf{42.33} & \textbf{50.45} & \textbf{61.34} & \textbf{16.80} & \textbf{22.20} & \textbf{25.20} & 27.26 & \textbf{31.58} & \textbf{39.03} & \textbf{37.02} & \textbf{42.81} & \textbf{52.6} & \textbf{42.44} & \textbf{51.74} & \textbf{3.05} \\ 

\hline

\multirow{2}{*}{BGE}
& unif & 40.32 & 43.76 & 57.29 & 12.46 & 14.54 & 17.85 & \textbf{32.27} & 36.14 & 51.04 & 27.71 & 33.49 & 49.6 & 32.54 & 35.52 & 4.20 \\
& lbl & \textbf{41.87} & \textbf{50.55} & \textbf{57.29} & \textbf{14.91} & \textbf{19.63} & \textbf{17.85} & 29.98 & \textbf{43.39} & \textbf{51.04} & \textbf{33.89} & \textbf{42.03} & \textbf{49.6} & \textbf{37.57} & \textbf{48.23} & \textbf{4.20} \\

\hline

\multirow{2}{*}{SF}

& unif & 40.53 & 45.19 & 61.46 & 13.47 & 15.79 & 16.54 & \textbf{31.94} & 30.93 & 45.05 & 23.59 & 27.56 & 66.2 & 41.07 & 42.22 & 2.04 \\
& lbl & \textbf{41.24} & \textbf{53.88} & \textbf{61.46} & \textbf{15.56} & \textbf{21.07} & \textbf{16.54} & 31.42 & \textbf{45.45} & \textbf{45.05} & \textbf{25.15} & \textbf{27.36} & \textbf{66.2} & \textbf{48.94} & \textbf{56.60} & \textbf{2.04} \\

\hline

\multirow{2}{*}{all-MPNet} 
& unif & 37.92 & 43.51 & 59.71 & 13.94 & 16.33 & 17.85 & 25.50 & 29.59 & 49.72 & 34.47 & 43.24 & 62.1 & 41.01 & 43.86 & 2.37 \\
& lbl & \textbf{38.37} & \textbf{49.57} & \textbf{59.71} & \textbf{21.42} & \textbf{22.87} & \textbf{17.85} & \textbf{27.82} & \textbf{37.54} & \textbf{49.72} & \textbf{37.20} & \textbf{42.10} & \textbf{62.1} & \textbf{53.53} & \textbf{60.10} & \textbf{2.37} \\ 

\hline

\multirow{2}{*}{all-MPNet-Wiki} 
& unif & 35.64 & 37.74 & 47.01 & 15.98 & 17.77 & 21.26 & 26.01 & 38.33 & 56.30 & 34.36 & 39.34 & 57.2 & 36.00 & 40.01 & 3.40 \\
& lbl & \textbf{37.11} & \textbf{44.22} & \textbf{47.01} & \textbf{12.16} & \textbf{26.58} & \textbf{21.26} & \textbf{26.48} & \textbf{48.28} & \textbf{56.30} & \textbf{36.38} & \textbf{46.51} & \textbf{57.2} & \textbf{42.14} & \textbf{53.73} & \textbf{3.40} \\ 

\hline

\multirow{2}{*}{GloVe} 
& unif & 33.85 & 35.90 & 19.70 & 10.56 & 11.56 & 7.35 & 6.73 & 2.94 & 0.00 & 9.14 & 9.68 & 4.7 & 33.70 & 32.45 & 8.82 \\
& lbl & \textbf{33.22} & \textbf{42.73} & \textbf{19.70} & \textbf{18.19} & \textbf{23.53} & \textbf{7.35} & \textbf{13.34} & \textbf{9.69} & \textbf{0.00} & \textbf{13.33} & \textbf{15.16} & \textbf{4.7} & \textbf{38.31} & \textbf{44.13} & \textbf{8.82}

\end{tabular}%
}
\end{table*}

\end{document} 

\bibliographystyle{unsrt}
\bibliography{references}  

\begin{thebibliography}{10}

\bibitem{tarekegn2024deeplearningmultilabellearning}
Adane~Nega Tarekegn, Mohib Ullah, and Faouzi~Alaya Cheikh.
\newblock Deep learning for multi-label learning: A comprehensive survey, 2024.

\bibitem{Isnaini_2019}
Nikmah Isnaini, Adiwijaya, Mohamad~Syahrul Mubarok, and Muhammad Yuslan~Abu Bakar.
\newblock A multi-label classification on topics of indonesian news using k-nearest neighbor.
\newblock {\em Journal of Physics: Conference Series}, 1192(1):012027, mar 2019.

\bibitem{axioms11090436}
Emre Deniz, Hasan Erbay, and Mustafa Coşar.
\newblock Multi-label classification of e-commerce customer reviews via machine learning.
\newblock {\em Axioms}, 11(9), 2022.

\bibitem{deng-fuji-2023-multilabel}
Jiawen Deng and Fuji Ren.
\newblock Multi-label emotion detection via emotion-specified feature extraction and emotion correlation learning.
\newblock {\em IEEE Transactions on Affective Computing}, 14(1):475--486, 2023.

\bibitem{huang-etal-2013-social}
Shu Huang, Wei Peng, Jingxuan Li, and Dongwon Lee.
\newblock Sentiment and topic analysis on social media: a multi-task multi-label classification approach.
\newblock In {\em Proceedings of the 5th Annual ACM Web Science Conference}, WebSci '13, page 172–181, New York, NY, USA, 2013. Association for Computing Machinery.

\bibitem{Lemmens_Dejaeghere_Kreutz_VanNooten_Markov_Daelemans_2021}
Jens Lemmens, Tess Dejaeghere, Tim Kreutz, Jens Van~Nooten, Ilia Markov, and Walter Daelemans.
\newblock Vaccinpraat: Monitoring vaccine skepticism in dutch twitter and facebook comments.
\newblock {\em Computational Linguistics in the Netherlands Journal}, 11:173–188, Dec. 2021.

\bibitem{vannootendaelemans2023improving}
Jens Van~Nooten and Walter Daelemans.
\newblock Improving {D}utch vaccine hesitancy monitoring via multi-label data augmentation with {GPT}-3.5.
\newblock In Jeremy Barnes, Orph{\'e}e De~Clercq, and Roman Klinger, editors, {\em Proceedings of the 13th Workshop on Computational Approaches to Subjectivity, Sentiment, {\&} Social Media Analysis}, pages 251--270, Toronto, Canada, July 2023. Association for Computational Linguistics.

\bibitem{van-olemn-etal-2022-predicting}
Josefien Van~Olmen, Jens Van~Nooten, Hilde Philips, Annet Sollie, and Walter Daelemans.
\newblock Predicting covid-19 symptoms from free text in medical records using artificial intelligence: Feasibility study.
\newblock {\em JMIR Med Inform}, 10(4):e37771, Apr 2022.

\bibitem{xiongyanmei2018subject}
Changzhen Xiong and Yanmei Shan.
\newblock Subject features and hash codes for multi-label image retrieval.
\newblock In {\em 2018 IEEE 7th Data Driven Control and Learning Systems Conference (DDCLS)}, pages 808--812, 2018.

\bibitem{wangetal2019baseline}
Qian Wang, Ning Jia, and Toby~P. Breckon.
\newblock A baseline for multi-label image classification using an ensemble of deep convolutional neural networks.
\newblock In {\em 2019 IEEE International Conference on Image Processing (ICIP)}, pages 644--648, 2019.

\bibitem{gupta2024openvocabularymultilabelvideo}
Rohit Gupta, Mamshad~Nayeem Rizve, Jayakrishnan Unnikrishnan, Ashish Tawari, Son Tran, Mubarak Shah, Benjamin Yao, and Trishul Chilimbi.
\newblock Open vocabulary multi-label video classification, 2024.

\bibitem{mulimani2024classincrementallearningmultilabelaudio}
Manjunath Mulimani and Annamaria Mesaros.
\newblock Class-incremental learning for multi-label audio classification, 2024.

\bibitem{forero-etal-2013-comparison}
A.~F. Giraldo-Forero, J.~A. Jaramillo-Garzón, and C.~G. Castellanos-Domínguez.
\newblock A comparison of multi-label techniques based on problem transformation for protein functional prediction.
\newblock In {\em 2013 35th Annual International Conference of the IEEE Engineering in Medicine and Biology Society (EMBC)}, pages 2688--2691, 2013.

\bibitem{szalkaigrolmusz2018protein}
Balázs Szalkai and Vince Grolmusz.
\newblock Near perfect protein multi-label classification with deep neural networks.
\newblock {\em Methods}, 132:50--56, 2018.
\newblock Comparison and Visualization Methods for High-Dimensional Biological Data.

\bibitem{chang2008importance}
Ming-Wei Chang, Lev Ratinov, Dan Roth, and Vivek Srikumar.
\newblock Importance of semantic representation: dataless classification.
\newblock In {\em Proceedings of the 23rd National Conference on Artificial Intelligence - Volume 2}, AAAI'08, page 830–835. AAAI Press, 2008.

\bibitem{Kosar-etal-2022}
Andriy Kosar, Guy De~Pauw, and Walter Daelemans.
\newblock Unsupervised text classification with neural word embeddings.
\newblock {\em Computational Linguistics in the Netherlands Journal}, 12:165–181, Dec. 2022.

\bibitem{kosar-etal-2023-advancing}
Andriy Kosar, Guy De~Pauw, and Walter Daelemans.
\newblock Advancing topical text classification: A novel distance-based method with contextual embeddings.
\newblock In Ruslan Mitkov and Galia Angelova, editors, {\em Proceedings of the 14th International Conference on Recent Advances in Natural Language Processing}, pages 586--597, Varna, Bulgaria, September 2023. INCOMA Ltd., Shoumen, Bulgaria.

\bibitem{veeranna2016using}
Sappadla~Prateek Veeranna, Jinseok Nam, Eneldo~Loza Menc{\i}a, and Johannes F{\"u}rnkranz.
\newblock Using semantic similarity for multi-label zero-shot classification of text documents.
\newblock In {\em Proceeding of european symposium on artificial neural networks, computational intelligence and machine learning. bruges, belgium: Elsevier}, pages 423--428, 2016.

\bibitem{mylonas-etal-2020-zero}
Nikolaos Mylonas, Stamatis Karlos, and Grigorios Tsoumakas.
\newblock Zero-shot classification of biomedical articles with emerging mesh descriptors.
\newblock In {\em 11th Hellenic Conference on Artificial Intelligence}, SETN 2020, page 175–184, New York, NY, USA, 2020. Association for Computing Machinery.

\bibitem{mustafa2021multi}
Ghulam Mustafa, Muhammad Usman, Lisu Yu, Muhammad~Tanvir Afzal, Muhammad Sulaiman, and Abdul Shahid.
\newblock Multi-label classification of research articles using word2vec and identification of similarity threshold.
\newblock {\em Scientific Reports}, 11(1):21900, 2021.

\bibitem{sarkaretal2023zero}
Souvika Sarkar, Dongji Feng, and Shubhra~Kanti Karmaker~Santu.
\newblock Zero-shot multi-label topic inference with sentence encoders and {LLM}s.
\newblock In Houda Bouamor, Juan Pino, and Kalika Bali, editors, {\em Proceedings of the 2023 Conference on Empirical Methods in Natural Language Processing}, pages 16218--16233, Singapore, December 2023. Association for Computational Linguistics.

\bibitem{sarkar-etal-2022-exploring}
Souvika Sarkar, Dongji Feng, and Shubhra~Kanti Karmaker~Santu.
\newblock {E}xploring universal sentence encoders for zero-shot text classification.
\newblock In Yulan He, Heng Ji, Sujian Li, Yang Liu, and Chua-Hui Chang, editors, {\em Proceedings of the 2nd Conference of the Asia-Pacific Chapter of the Association for Computational Linguistics and the 12th International Joint Conference on Natural Language Processing (Volume 2: Short Papers)}, pages 135--147, Online only, November 2022. Association for Computational Linguistics.

\bibitem{zhang-etal-2018-binaryrelevance}
Min-Ling Zhang, Yu-Kun Li, Xu-Ying Liu, and Xin Geng.
\newblock Binary relevance for multi-label learning: an overview.
\newblock {\em Frontiers of Computer Science}, 12(2):191--202, 2018.

\bibitem{fan-qiu-2023-hierarchical}
Qingwu Fan and Changsheng Qiu.
\newblock Hierarchical multi-label text classification method based on multi-level decoupling.
\newblock In {\em 2023 3rd International Conference on Neural Networks, Information and Communication Engineering (NNICE)}, pages 453--457, 2023.

\bibitem{jia-etal-2022-research}
Ling Jia, Jin Fan, Dong Sun, Qingwei Gao, and Yixiang Lu.
\newblock Research on multi-label classification problems based on neural networks and label correlation.
\newblock In {\em 2022 41st Chinese Control Conference (CCC)}, pages 7298--7302, 2022.

\bibitem{cai-etal-2023-resolving}
Xunxin Cai, Meng Xiao, Zhiyuan Ning, and Yuanchun Zhou.
\newblock Resolving the imbalance issue in hierarchical disciplinary topic inference via llm-based data augmentation.
\newblock In {\em 2023 IEEE International Conference on Data Mining Workshops (ICDMW)}, pages 1424--1429, 2023.

\bibitem{benbaruch2021asymmetricloss}
Emanuel Ben-Baruch, Tal Ridnik, Nadav Zamir, Asaf Noy, Itamar Friedman, Matan Protter, and Lihi Zelnik-Manor.
\newblock Asymmetric loss for multi-label classification, 2021.

\bibitem{li-etal-2022-survey}
Qian Li, Hao Peng, Jianxin Li, Congying Xia, Renyu Yang, Lichao Sun, Philip~S. Yu, and Lifang He.
\newblock A survey on text classification: From traditional to deep learning.
\newblock {\em ACM Trans. Intell. Syst. Technol.}, 13(2), apr 2022.

\bibitem{devlin-etal-2019-bert}
Jacob Devlin, Ming-Wei Chang, Kenton Lee, and Kristina Toutanova.
\newblock {BERT}: Pre-training of deep bidirectional transformers for language understanding.
\newblock In Jill Burstein, Christy Doran, and Thamar Solorio, editors, {\em Proceedings of the 2019 Conference of the North {A}merican Chapter of the Association for Computational Linguistics: Human Language Technologies, Volume 1 (Long and Short Papers)}, pages 4171--4186, Minneapolis, Minnesota, June 2019. Association for Computational Linguistics.

\bibitem{Radford2018ImprovingLU}
Alec Radford and Karthik Narasimhan.
\newblock Improving language understanding by generative pre-training.
\newblock In {\em arxiv}, 2018.

\bibitem{yin2024crisissensellminstructionfinetunedlarge}
Kai Yin, Chengkai Liu, Ali Mostafavi, and Xia Hu.
\newblock Crisissense-llm: Instruction fine-tuned large language model for multi-label social media text classification in disaster informatics, 2024.

\bibitem{malik-etal-2024-pseudo}
Usman Malik, Simon Bernard, Alexandre Pauchet, Clément Chatelain, Romain Picot-Clémente, and Jérôme Cortinovis.
\newblock Pseudo-labeling with large language models for multi-label emotion classification of french tweets.
\newblock {\em IEEE Access}, 12:15902--15916, 2024.

\bibitem{van-nooten-kosar-2024-advancing}
Jens Van~Nooten and Andriy Kosar.
\newblock Advancing {CSR} theme and topic classification: {LLM}s and training enhancement insights.
\newblock In Chung-Chi Chen, Xiaomo Liu, Udo Hahn, Armineh Nourbakhsh, Zhiqiang Ma, Charese Smiley, Veronique Hoste, Sanjiv~Ranjan Das, Manling Li, Mohammad Ghassemi, Hen-Hsen Huang, Hiroya Takamura, and Hsin-Hsi Chen, editors, {\em Proceedings of the Joint Workshop of the 7th Financial Technology and Natural Language Processing, the 5th Knowledge Discovery from Unstructured Data in Financial Services, and the 4th Workshop on Economics and Natural Language Processing}, pages 292--305, Torino, Italia, May 2024. Association for Computational Linguistics.

\bibitem{wang-etal-2023-text2topic}
Fengjun Wang, Moran Beladev, Ofri Kleinfeld, Elina Frayerman, Tal Shachar, Eran Fainman, Karen Lastmann~Assaraf, Sarai Mizrachi, and Benjamin Wang.
\newblock {T}ext2{T}opic: Multi-label text classification system for efficient topic detection in user generated content with zero-shot capabilities.
\newblock In Mingxuan Wang and Imed Zitouni, editors, {\em Proceedings of the 2023 Conference on Empirical Methods in Natural Language Processing: Industry Track}, pages 93--103, Singapore, December 2023. Association for Computational Linguistics.

\bibitem{BOGATINOVSKI2022117215}
Jasmin Bogatinovski, Ljupčo Todorovski, Sašo Džeroski, and Dragi Kocev.
\newblock Comprehensive comparative study of multi-label classification methods.
\newblock {\em Expert Systems with Applications}, 203:117215, 2022.

\bibitem{pennington-etal-2014-glove}
Jeffrey Pennington, Richard Socher, and Christopher Manning.
\newblock {G}lo{V}e: Global vectors for word representation.
\newblock In Alessandro Moschitti, Bo~Pang, and Walter Daelemans, editors, {\em Proceedings of the 2014 Conference on Empirical Methods in Natural Language Processing ({EMNLP})}, pages 1532--1543, Doha, Qatar, October 2014. Association for Computational Linguistics.

\bibitem{word2vec}
Tomas Mikolov, Kai Chen, Greg Corrado, and Jeffrey Dean.
\newblock Efficient estimation of word representations in vector space, 2013.

\bibitem{fasttext}
Piotr Bojanowski, Edouard Grave, Armand Joulin, and Tomas Mikolov.
\newblock Enriching word vectors with subword information, 2017.

\bibitem{reimers2019sentence-bert}
Nils Reimers and Iryna Gurevych.
\newblock Sentence-bert: Sentence embeddings using siamese bert-networks.
\newblock In {\em Proceedings of the 2019 Conference on Empirical Methods in Natural Language Processing}. Association for Computational Linguistics, 11 2019.

\bibitem{radeva-etal-2024-similarity}
Irina Radeva, Ivan Popchev, and Miroslava Dimitrova.
\newblock Similarity thresholds in retrieval-augmented generation.
\newblock In {\em 2024 IEEE 12th International Conference on Intelligent Systems (IS)}, pages 1--7, 2024.

\bibitem{mohammad-etal-2018-semeval}
Saif Mohammad, Felipe Bravo-Marquez, Mohammad Salameh, and Svetlana Kiritchenko.
\newblock {S}em{E}val-2018 task 1: Affect in tweets.
\newblock In Marianna Apidianaki, Saif~M. Mohammad, Jonathan May, Ekaterina Shutova, Steven Bethard, and Marine Carpuat, editors, {\em Proceedings of the 12th International Workshop on Semantic Evaluation}, pages 1--17, New Orleans, Louisiana, June 2018. Association for Computational Linguistics.

\bibitem{apte-etal-reuters}
Chidanand Apt{'{e}}, Fred Damerau, and Sholom~M. Weiss.
\newblock Automated learning of decision rules for text categorization.
\newblock {\em ACM Transactions on Information Systems}, 1994.
\newblock To appear.

\bibitem{yang-etal-2018-sgm}
Pengcheng Yang, Xu~Sun, Wei Li, Shuming Ma, Wei Wu, and Houfeng Wang.
\newblock {SGM}: Sequence generation model for multi-label classification.
\newblock In Emily~M. Bender, Leon Derczynski, and Pierre Isabelle, editors, {\em Proceedings of the 27th International Conference on Computational Linguistics}, pages 3915--3926, Santa Fe, New Mexico, USA, August 2018. Association for Computational Linguistics.

\bibitem{chen-etal-litcovid}
Qingyu Chen, Alexis Allot, and Zhiyong Lu.
\newblock {LitCovid: an open database of COVID-19 literature}.
\newblock {\em Nucleic Acids Research}, 49(D1):D1534--D1540, 11 2020.

\bibitem{mlstratification}
Konstantinos Sechidis, Grigorios Tsoumakas, and Ioannis Vlahavas.
\newblock On the stratification of multi-label data.
\newblock In Dimitrios Gunopulos, Thomas Hofmann, Donato Malerba, and Michalis Vazirgiannis, editors, {\em Machine Learning and Knowledge Discovery in Databases}, pages 145--158, Berlin, Heidelberg, 2011. Springer Berlin Heidelberg.

\bibitem{solatorio2024gistembed}
Aivin~V. Solatorio.
\newblock Gistembed: Guided in-sample selection of training negatives for text embedding fine-tuning.
\newblock {\em arXiv preprint arXiv:2402.16829}, 2024.

\bibitem{li2023towards}
Zehan Li, Xin Zhang, Yanzhao Zhang, Dingkun Long, Pengjun Xie, and Meishan Zhang.
\newblock Towards general text embeddings with multi-stage contrastive learning.
\newblock {\em arXiv preprint arXiv:2308.03281}, 2023.

\bibitem{li2023angle}
Xianming Li and Jing Li.
\newblock Angle-optimized text embeddings.
\newblock {\em arXiv preprint arXiv:2309.12871}, 2023.

\bibitem{emb2024mxbai}
Sean Lee, Aamir Shakir, Darius Koenig, and Julius Lipp.
\newblock Open source strikes bread - new fluffy embeddings model, 2024.

\bibitem{liu-etal-2019-roberta}
Yinhan Liu, Myle Ott, Naman Goyal, Jingfei Du, Mandar Joshi, Danqi Chen, Omer Levy, Mike Lewis, Luke Zettlemoyer, and Veselin Stoyanov.
\newblock Roberta: A robustly optimized bert pretraining approach, 2019.

\bibitem{geminiteam2024gemini15unlockingmultimodal}
Gemini Team, Petko Georgiev, Ving~Ian Lei, Ryan Burnell, Libin Bai, Anmol Gulati, and Garrett Tanzer.
\newblock Gemini 1.5: Unlocking multimodal understanding across millions of tokens of context, 2024.

\bibitem{zhou2022problemscosinemeasureembedding}
Kaitlyn Zhou, Kawin Ethayarajh, Dallas Card, and Dan Jurafsky.
\newblock Problems with cosine as a measure of embedding similarity for high frequency words, 2022.

\bibitem{mlcm}
Mohammadreza Heydarian, Thomas~E. Doyle, and Reza Samavi.
\newblock Mlcm: Multi-label confusion matrix.
\newblock {\em IEEE Access}, 10:19083--19095, 2022.

\end{thebibliography}






\end{document}